%% file: iclr2026_conference.tex
\documentclass{article} % For LaTeX2e
\usepackage{iclr2026_conference,times}

\input{math_commands.tex}

\usepackage{hyperref}
\usepackage{url}

\usepackage{bm}
\usepackage{enumitem}
\usepackage{amsmath}
\usepackage{multirow}
\usepackage{multicol}
\usepackage{graphicx}
\usepackage[table,xcdraw]{xcolor}
\usepackage{tabularx}
\usepackage{booktabs}
\usepackage{soul}

\definecolor{mygray}{HTML}{7f7f7f}   
\definecolor{mygreen}{HTML}{4fad5b}

\title{Task-Related Token Compression in Multimodal Large Language Models from an Explainability Perspective}

\author{
    \And   Lei Lei$^{1,3}$\thanks{Work done during interships in Rightly Robotics.}    
    \quad   Jie Gu$^{2}$\thanks{Corresponding author.}
    \quad   Xiaokang Ma$^{2}$
    \quad   Chu Tang$^{2}$   
    \quad   Jingmin Chen$^{2}$
    \quad   Tong Xu$^{1}$\footnotemark[2]\\  
    $^1$ University of Science and Technology of China \quad 
    $^2$ Rightly Robotics \\ 
    $^3$ Shanghai Innovation Institute \\
    {\small \texttt{\{lily168,tongxu\}@ustc.edu.cn}; \texttt{\{jgu,xma,chu.tang,jingmin.chen\}@rightly.ai}}
}

\usepackage{titlesec}

\iclrfinalcopy
\begin{document}

\maketitle
\input{chapters/0.abstract}
\input{chapters/1.introduction}
\input{chapters/2.relatedwork}
\input{chapters/3.method}
\input{chapters/4.experiments}
\input{chapters/5.conclusion}

\noindent\textbf{Reproducibility Statement.} We provide the necessary information to facilitate reproducibility. Experimental settings and implementation details are described in the Section~\ref{sec:4_1} and Appendix~\ref{app:2}. All the datasets used in our experiments are publicly available, and the preprocessing steps of training data are documented in Appendix~\ref{app:2}.

\section*{Acknowledgements}
This work was supported in part by the grants from National Science and Technology Major Project (No. 2023ZD0121104), and National Natural Science Foundation of China (No.62222213, U22B2059)

\bibliography{iclr2026_conference}
\bibliographystyle{iclr2026_conference}

\newpage
\input{chapters/appendix}

\end{document}

%% file: math_commands.tex
\usepackage{amsmath,amsfonts,bm}

\def\eqref#1{equation~\ref{#1}}
\def\1{\bm{1}}

\DeclareMathAlphabet{\mathsfit}{\encodingdefault}{\sfdefault}{m}{sl}
\SetMathAlphabet{\mathsfit}{bold}{\encodingdefault}{\sfdefault}{bx}{n}

%% file: chapters/0.abstract.tex
\begin{abstract}
% Existing Large Vision Models (LVMs) typically process a large number of visual tokens that contain significant redundancy, especially when handling videos. In this work, we focus on LVMs for video and explore an under-investigated topic: the direct removal of redundancy at the token level before feeding these tokens into the large language model (LLM). Specifically, token compression can be achieved by leveraging explainability methods to identify the visual signals that LVMs most attend to. Our evaluations with prominent models such as QwenVL and LLaVA demonstrate the effectiveness of this explainability-based token compressor. Furthermore, we propose learning a mapping from the attention map of the first layer in LLM to the explainability results, enabling token compression to be completed prior to LLM computation and thus facilitating practical deployment. Interestingly, this can be achieved using a naive convolutional network, whose training is efficient and independent of LVMs. Extensive experiments demonstrate the ability of our approach to dramatically reduce computational costs (\emph{e.g.}, a 50\% reduction in visual tokens) without sacrificing performance. It also exhibits robust generalization even when the number of tokens in inference far exceeds that used in training. We believe that this work not only offers practical value for efficient deployment, but also provides valuable and universally applicable insights.

Existing Multimodal Large Language Models (MLLMs) process a large number of visual tokens, leading to significant computational costs and inefficiency. Instruction-related visual token compression demonstrates strong task relevance, which aligns well with MLLMs’ ultimate goal of instruction following. Previous works generally assume that visual tokens achieve better vision–language alignment in the shallow layers of LLMs, which have led to task-related token compression being primarily applied in intermediate LLM layers. In contrast, our study reveals that with proper selection, task-related token compression is feasible at the input stage of LLM with negligible performance loss. This new paradigm significantly reduces task-irrelevant visual tokens and its model-agnostic design enables application without modifying the LLM architecture. Specifically, we suggest that explainability methods for transformer-based architechtures can evaluate the global importance of each visual token with respect to the given instruction, which can effectively guide the task-related token compression for MLLMs. Furthermore, we propose to learn a mapping from the attention map of the first LLM layer to the explanation results, thereby avoiding the need for a full inference pass. Interestingly, this mapping can be learned using a simple and lightweight convolutional network, whose training is efficient and independent of MLLMs. Extensive experiments on $13$ image and video benchmarks across three leading MLLMs (Qwen2-VL, LLaVA-OneVision, and VILA1.5) demonstrate the remarkable effectiveness and strong generalization of our approach. Additionally, our new compression paradigm achieves faster inference with reductions in both prefilling time and KV-cache memory.

% It also exhibits strong generalization, even when the number of tokens in inference far exceeds that used in training. 

  % The abstract paragraph should be indented \nicefrac{1}{2}~inch (3~picas) on
  % both the left- and right-hand margins. Use 10~point type, with a vertical
  % spacing (leading) of 11~points.  The word \textbf{Abstract} must be centered,
  % bold, and in point size 12. Two line spaces precede the abstract. The abstract
  % must be limited to one paragraph.
\end{abstract}

%% file: chapters/1.introduction.tex
\section{Introduction}
With large language models (LLMs) providing a strong foundation~\cite{llm20_arxiv, gpt4_arxiv, llama_arxiv, deepseek_arxiv}, research on multimodal large language models (MLLMs) has gained significant momentum~\cite{llava_nips, internvl_arxiv, minigpt4_iclr, qwenvl_arxiv}. Considerable progress has been achieved in various image- and video-related tasks~\cite{gpt4v_arxiv, gemini_arxiv}. A common paradigm among existing MLLMs is to jointly feed visual tokens (generated by a vision encoder) and textual tokens into the LLM for cross-modal alignment and integration~\cite{llava_nips, minigpt4_iclr, BLIP2}. This paradigm introduces substantial memory and computational overhead due to the high volume of visual tokens, which grows rapidly with higher resolutions or frame rates~\cite{qwen2vl_arxiv, internlmX2.5_arxiv}. Consequently, there is a pressing need for effective token compression techniques.
% Large Vision Models (LVMs) have made substantial progress with strong performance across a range of multi-modal tasks. Large Language Model (LLM), as the backbone of LVM, enable the integration and understanding of visual and textual information, playing a key role in multi-modal reasoning. However, visual information is typically sparse. The inherent redundancy in visual tokens causes excessive memory and computational overhead within LLM. This issue is particularly pronounced in video scenarios, where the volume of visual tokens can grow rapidly with higher resolution or frame rates. Consequently, there is a pressing need for effective token compression to reduce visual redundancy.

Previous exploration of visual token compression methods can be roughly divided into two categories. The first aims to obtain more compact and fewer visual representations (especially for videos) in a task/instruction-agnostic manner (independent of LLM)~\cite{tome_iclr, visionzip_arxiv, tempme_iclr, folder_arxiv, fastvid_arxiv}. We argue that visual representations are an integral part of MLLMs and serve as the foundation for achieving strong performance and generalization. Many state-of-the-art(SOTA) MLLMs already incorporate built-in, task-agnostic compression mechanisms (e.g., spatial and temporal pooling) instead of relying on separate compression techniques applied afterward~\cite{qwen2vl_arxiv,Dynamic-VLM,LLAMA-vid}. The second category focuses on selecting visual tokens that are most relevant to the given instruction. FastV~\cite{fastv_eccv} is a pioneering work that highlights the importance of retaining all shallow-layer visual tokens in LLMs for lossless compression. While this assumption has been adopted by many subsequent studies~\cite{sparsevlm_arxiv, smallforlarge_arxiv, pdrop_arxiv, tokencarve_arxiv, prunevid_arxiv, Dart}, we believe it remains open to question: \textit{Are all visual tokens in the shallow layers of LLM truly indispensable for task-related compression?}

% \textcolor{blue}{Many subsequent works follow this paradigm to perform intermediate-layer compression. However, the importance of tokens is determined by intermediate representations, which lack global information. Consequently, some tokens that receive higher attention in deeper layers may be removed prematurely in shallow layers, leading to performance degradation. Moreover, compression in intermediate layers require modifications to the original LLM architecture and introduce a burden of hyperparameter tuning, e.g., compression layer selection.}

% Recent work primarily explore token compression during the prefilling stage. FastV proposes a training-free token compression strategy based on statistical observations from a specific model, leveraging shallow-layer attention maps to identify and remove less important visual tokens. Many subsequent works derive various compression strategies based on attention maps selected in LLM, focusing on prefilling-stage token compression. Despite its effectiveness, compressing tokens in intermediate layers of the LLM relies on insights derived from specific architectures, limiting its generality and transferability across architectures. Additionally, in practical deployment, prefilling-stage token compression still requires storing the KV cache of all visual tokens, including those that are eventually removed, leading to limited memory savings. Moreover, we prefer not to impose structural constraints on LVMs that would require it to be compatible with various compression strategies.

This paper seeks to answer the question of whether an effective task-related token compression approach prior to the LLM exists but remains undiscovered, or whether it is inherently infeasible. To the best of our knowledge, our work is among the earliest efforts to investigate this issue. We first explore the use of explainability methods to assess visual token importance with respect to the instruction. Explainability methods for transformer-based architecture generally iteratively update a relevance map across layers using gradient-weighted multi-head attentions~\cite{gae_iccv,CheferGW21}, which effectively captures the global relevance scores of visual tokens. Relevance scores indicating the contributions of input tokens to output can be used to rank and prune less important visual tokens for compression.
% In this work, we focus on token-level compression during the input stage, an under-investigated topic with great potential for reducing visual redundancy. Regarding how to obtain token-level importance, explainability methods offer a practical solution by identifying relevance distributions from output tokens to input tokens. Specifically, explainability methods for transformer-based architecture initializes a relevance map as an identity matrix and iteratively updates it across layers using gradient-weighted multi-head attentions. Afterward, visual relevance scores can be extracted from the resulting relevance map as importance indicator to rank and remove less important visual tokens at the input stage. We evaluate this explainability-based token compression approach on QwenVL and LLaVA with video inputs. Experiments show that the method reduces the number of visual tokens by 50\% while maintaining over 99\% of the original accuracy. This highlights the strong capability of explainability methods to assess visual token importance and further guide input-stage token compression.
Comprehensive experiments conducted on both image and video data across three representative MLLMs demonstrate the effectiveness of such a compressor. The results indicate that, with proper selection, pruning tokens that are not relevant to the task at the LLM input stage is indeed feasible. Moreover, unlike previous works motivated by observations derived from specific network architectures (\emph{e.g.}, LLaVA)~\cite{fastv_eccv, tokencarve_arxiv}, which limits their generality and transferability, our explainability-based approach is broadly applicable. Rather than relying on the behaviors of specific models, it leverages the inherent characteristics of the applied model.

After validating that the explanation results are effective compression indicators, a lightweight model capable of generating an alternative to the relevance map is further needed to enable efficient and practical deployment. Interestingly, this goal can be achieved by training a simple fully convolutional network that predicts relevance based on the first-layer attention map of the LLM. The training process is highly efficient (\emph{e.g.}, training a 5-layer network using only 10K image data) and does not involve any changes to the MLLM itself. Using the predicted relevance, token compression can be performed prior to the prefill phase with negligible extra computational cost. As a result, both computational and memory overhead during inference are significantly reduced, with no modifications required to either the prefill or decode phases. Last but not least, our approach generalizes well across various architectures, benefiting from the broadly applicable nature of explainability methods and the MLLM-agnostic design of the auxiliary training.
% However, the implementation of explainability methods requires a full inference pass, including both forward and backward propagation, which renders it infeasible for deployment in token compression. Since the relevance map is essentially an aggregation of attention maps, we propose to learn a mapping from the first-layer attention map to the final relevance map ---
% a trade-off between efficiency and effectiveness. Notably, a naive 1D convolutional network is sufficient to take on this task. This provides a lightweight and scalable alternative for obtaining relevance maps, addressing the practical limitations of explainability-based token compression.

% Building on these insights, we introduce an explainability-based token-level compression method for VLMs. Given a VLM, we propose to learn its attention patterns over visual information using a lightweight convolutional network. The network is trained on data generated via explainability methods, which provide token-level relevance distribution by tracing the contributions from output tokens back to input visual tokens. Once trained, the convolutional model is deployed at the input stage to map first-layer attention scores into token importance rankings, enabling effective token compression. This approach is generic --- it generalizes well across architectures, as explainability methods are broadly applicable and the auxiliary training is VLM-agnostic. Furthermore, it is highly deployable, with negligible training cost and minimal inference overhead, while maintaining strong performance.

To thoroughly assess the capability of our approach, we apply it to three prominent models with different architectures and visual representations: VILA1.5, LLaVA-OneVision, and Qwen2-VL. We include 13 widely used image and video benchmarks that span a wide range of visual complexities and tasks, ensuring a comprehensive evaluation. Notably, our method achieves significant compression by pruning 75\% of video tokens while retaining more than 97\% of the original performance across all benchmarks for both VILA1.5 and LLaVA-OneVision. It also performs well on image tasks, where up to 50\% of image tokens can be removed with only a minimal performance drop: maintaining over 96\% of baseline performance for Qwen2-VL and LLaVA-OneVision.

% We conduct extensive experiments on 10 widely used image and video benchmarks. For image tasks, we adopt benchmarks such as MME, MMStar, MMVet and SEED-Bench. For video tasks, we choose benchmarks including Video-MME(wo sub.), MVBench, MMBench-Video, NExT-QA(multi-choice QA and open-ended QA) and ActivityNetQA. These benchmarks encompass various task types (e.g., multi-choice and open-ended), ensuring a comprehensive evaluation. 
%分析结果 In addition, 扩展结果%

% In summary, our contributions are as follows:
% \begin{itemize}[left=0.5em]
%     \item We utilize relevance map derived from explainability method to guide visual token compression, validating its capacity to accurately indicate visual tokens importance.
%     \item We introduce an explainability-based token-level compression method for VLMs. A convolutional network is independently trained to model the VLM’s attention patterns on visual information. This lightweight network is subsequently employed during the input stage for removing redundancy at the token level directly.
%     \item We conduct extensive experiments to demonstrate the effectiveness and generalizability of our method.
% \end{itemize}

In summary, the contributions of the work are threefold: (i) reveal that explainability methods can well evaluate the importance of visual tokens, enabling effective token compression. (ii) propose a highly efficient token compressor by learning from explanation results. It allows token compression to be performed before the LLM, significantly reducing inference costs at both the prefill and decode phases. (iii) Validate the effectiveness and generalization of our method through extensive experiments on a wide range of image and video benchmarks across different leading MLLMs.

% In summary, we make three key contributions in this work: (i) We utilize relevance map derived from explainability method to guide visual token compression, validating its capacity to accurately indicate visual tokens importance. (ii) We introduce an explainability-based token-level compression method for VLMs. A convolutional network is independently trained to model the VLM’s attention patterns on visual information. This lightweight network is subsequently employed during the input stage for removing redundancy at the token level directly. (iii) We conduct extensive experiments to demonstrate the effectiveness and generalizability of our method.

%% file: chapters/2.relatedwork.tex
\section{Related Work}

\noindent\textbf{Multimodal Large Language Models.} Benefiting from advancements in large language models (LLMs)~\cite{gpt4_arxiv, llama_arxiv, deepseek_arxiv}, multimodal large language models (MLLMs) have gained considerable attention due to the powerful ability in multi-modal understanding and reasoning~\cite{llava_nips, internvl_arxiv, qwenvl_arxiv, gpt4v_arxiv, gemini_arxiv}. Recent advances~\cite{llavaov_arxiv, qwen2vl_arxiv, internlmX2.5_arxiv} tend to handle images with higher resolution and videos with more frames, which significantly increases the number of visual tokens and thus the computational burden. This reveals the necessity for token compression strategies that can balance efficiency and effectiveness. Our work proposes a new token compression paradigm, which removes task-irrelevant visual tokens at the LLM input stage, significantly reducing computational costs without sacrificing performance.

\noindent\textbf{Visual Token Compression.} Existing visual token compression methods for MLLMs can be broadly categorized into: task/instruction-agnostic compression~\cite{tome_iclr, visionzip_arxiv, tempme_iclr, DivPrune} and task/instruction-related compression~\cite{fastv_eccv, pdrop_arxiv, prunevid_arxiv, Dart}. 
The first category of methods typically introduces additional modules to merge redundant visual tokens based on similarity, addressing the limitations of existing models. However, many recent works have developed techniques to obtain more compact visual representations when building MLLMs~\cite{qwen2vl_arxiv, efficientlmm_nips24}. Notably, task/instruction-related compression can further reduce the number of visual tokens on models already incorporate built-in compression mechanisms, offering greater potential for efficiency gains. FastV~\cite{fastv_eccv} represents a typical method of the second category, which rely on shallow-layer attention maps of the LLM for compression. In this work, we explores the feasibility of an effective task-related token compression prior to the LLM, which functions independently of the architecture and can be applied broadly across different MLLMs.
% The first category of methods typically merges redundant visual tokens based on similarity. 

%% file: chapters/3.method.tex
\begin{figure*}[t]
  \centering
  \includegraphics[width=\textwidth]{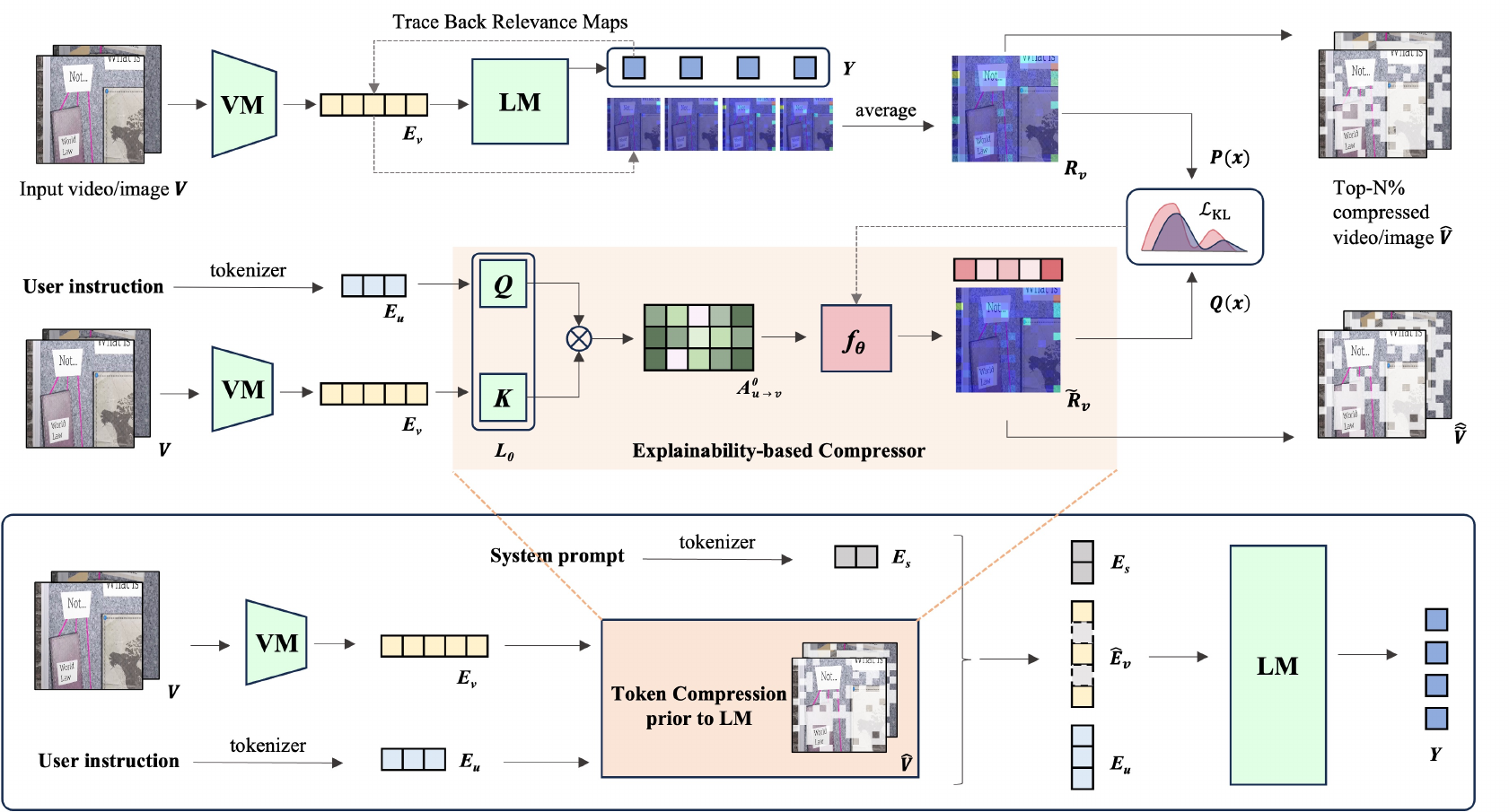}
  \caption{\textbf{Overview of our method.} The top portion illustrates the details of our explainability-based~compression approach: an explainability method can reveal the important visual tokens (first row, Section~\ref{sec:3_2}); a lightweight model can then be trained to approximate this explainability and serve as a compression indicator (second row, Section~\ref{sec:3_3}). The bottom portion shows a general inference framework for MLLMs, where the resulting compressor is applied at the LLM input stage.}
  \label{fig:overview}
\end{figure*}

\vspace{-0.2cm}

\section{Method}
\subsection{Background and Motivation}
\label{sec:3_1}
Current Multimodal Large Language Models (MLLMs) typically follow a framework in which a vision encoder is incorporated to encode visual signals into a sequence of tokens~\cite{llava_nips, qwenvl_arxiv, internvl_arxiv}. Specifically, multiple frames or patches are sampled from a video or an image, and their corresponding visual tokens are encoded. These visual tokens are then flattened and concatenated with textual prompt tokens before being fed into a Large Language Model (LLM) to generate a response. Formally, let $V$ be the video or image, and let $\mathbf{VM}$ and $\mathbf{LM}$ represent the vision encoder and the language model, respectively. The visual token embeddings $E_v$ can be represented as $E_v = \mathbf{VM}(V) \in \mathbb{R}^{N_v \times C}$, where $N_v$ is the number of visual tokens and $C$ is the feature dimension. \footnote{A cross-modal projector is commonly employed in such architectures. For notational simplicity, we denote both the vision encoder and the projector by $\mathbf{VM}$.} Let $E_s \in \mathbb{R}^{N_s \times C}$ and $E_u \in \mathbb{R}^{N_u \times C}$ denote the token embeddings of the system prompt and user instruction, respectively. By feeding $E_v$ together with $E_s$ and $E_u$ into the LLM, a textual response is generated, \emph{i.e.}, $Y = \mathbf{LM}(E_s, E_v, E_u)$. An additional compression module $\mathbf{Comp}$ can be introduced to prune visual tokens, while keeping the MLLM architecture—including both the $\mathbf{VM}$ and the $\mathbf{LM}$—unchanged during this pruning process.
% When processing videos, LVMs typically follow a framework where multiple frames are sampled from the input video and a vision encoder converts these frames into visual tokens. These tokens are then concatenated with textual prompt tokens and fed into a LLM to generate responses. Formally, given a video $V$, a frame sampler extracts $T$ frames from $V$ based on FPS sampling or uniform sampling. Then the video frames $F$ are encoded by a vision encoder model $\mathbf{VM}$ into a sequence of visual feature $E_v = \mathbf{VM}(F) \in \mathbb{R}^{N_v \times C}$, where $N_v$ represents the number of visual tokens and C the feature dimension. Specifically, each of the $T$ sampled frames is divided into a grid of $H \times W$ patches, resulting in $H \times W$ per frame. Therefore, the total number of visual tokens is given by $N_v = T \times H \times W$. Afterwards, the vision feature $E_v$ is processed together with the textual features $E_s \in \mathbb{R}^{N_s \times C}$ and $E_u \in \mathbb{R}^{N_u \times C}$, which represent the system prompt and user instruction respectively, by a language model $\mathbf{LM}$ to generate the textual response $Y = \mathbf{LM}(E_s, E_v, E_u)$.

$E_v$ can be considered as general-purpose representations of visual signals that are task/instruction-agnostic. Recent advances have already incorporated built-in compression mechanisms (e.g., spatial and temporal pooling) to reduce task-agnostic redundancy, enabling the construction of more compact $E_v$ when building MLLMs~\cite{qwen2vl_arxiv,Dynamic-VLM,LLAMA-vid}. Rather than applying another round of task-agnostic compression to $E_v$ (as in methods~\cite{tome_iclr, fastvid_arxiv}), our objective is to further address task-related redundancy by evaluating the importance of each token in $E_v$ with respect to a given instruction and selectively pruning those that contribute less. Recent advances have developed techniques to reduce the number of visual tokens to obtain a more compact $E_v$ when building MLLMs~\cite{qwen2vl_arxiv,Dynamic-VLM,LLAMA-vid}. Therefore, instead of further compressing $E_v$ in isolation (namely task-agnostic compresssion methods like~\cite{tome_iclr, fastvid_arxiv}), our objective is to assess the importance of each token in $E_v$ with respect to a given instruction, and subsequently prune those that are less essential. Moreover, we investigate how to perform token compression prior to LLM computation, \emph{i.e.}, compressing $E_v$ to $\hat{E_v} = \mathbf{Comp}(E_v | E_u) \in \mathbb{R}^{\hat{N}_v \times C}$ and then computing $Y = \mathbf{LM}(E_s, \hat{E_v}, E_u)$, where $\hat{N}_v$ is much smaller than $N_v$. In contrast to previous methods~\cite{fastv_eccv, prunevid_arxiv}, our method does not require any modifications to the prefill and decode phases during inference, and computational and memory overhead can be significantly reduced in both phases. 

% This work encourages the research community to reconsider an intriguing question: \textit{are all visual tokens in the shallow layers of LLM truly essential?}

The details of our approach are presented below. In Section~\ref{sec:3_2}, we introduce explainability methods to assess the importance of visual tokens and guide token compression. A learning mechanism is then proposed to predict the explanation results in Section~\ref{sec:3_3}, which ultimately enables effective token compression at the LLM input stage.
% Our work focuses on input-stage token compression, enhancing the pipeline by using explainability cues to remove visual redundancy and generate refined feature $\hat{E_v}$ for subsequent language model processing. The compressed visual feature $\hat{E_v} \in \mathbb{R}^{\hat{N}_v} \times C$ contains fewer tokens than the original $E_v$, while preserving more relevant visual information.
% Therefore, the final output is formulated as $Y = \mathbf{LM}(E_s, \hat{E_v}, E_u)$.

\begin{figure*}
  \centering
  \includegraphics[width=\textwidth]{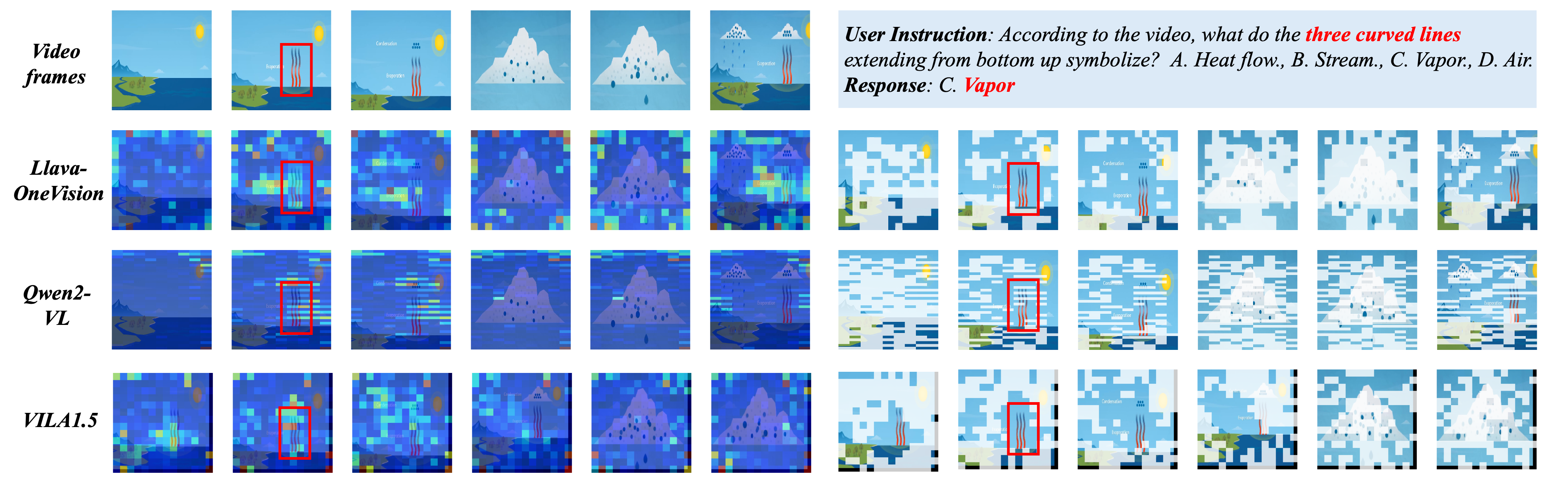}
  \caption{ \textbf{Visualization of $R_v$ obtained via the explainability method (left) and the corresponding token pruning results (right).} Based on $R_v$, the top 50\% of visual tokens are retained and the rest masked in white. Given the instruction querying the ``three curved lines", $R_v$ highlights the regions corresponding to the ``three curved lines", guiding the selective retention of the associated visual tokens. All three MLLMs generate the correct answer using only the retained tokens. More visualization cases are presented in Appendix~\ref{app:1}.}
  \label{fig:visualization}
\end{figure*}

\subsection{Token Compression with Explainability}
\label{sec:3_2}
To reduce task-related redundancy at the token level, we need to estimate the contribution of each visual token to the model response. Explainability methods for LLMs facilitate this goal by generating a relevance map through the integration of attention weights and corresponding gradients, effectively revealing where the model genuinely focuses. The resulting relevance map highlights the contributions, enabling us to rank and prune these visual tokens accordingly. The pipeline for this section is shown in the first row of Figure~\ref{fig:overview}.
% To remove redundancy at the token level, we require an estimate of the importance of each visual token. Explainability methods for LVMs facilitate this by generating relevance a map through the integration of attention weights and corresponding gradients, effectively revealing the model’s true focus. Based on the relevance map derived from explainability methods, we can rank visual tokens by their importance, enabling token-level compression at the input stage.

\noindent\textbf{Relevance Maps by Explainability Method.} We adopt a generic explainability method similar to ~\cite{deco_arxiv,gae_iccv} to compute the relevance of the response-to-vision. The relevance values reveal the~distribution of importance across visual tokens utilized by the LLM. Without loss of generality, assume that the LLM in an MLLM has $L$ layers, and denote the generated sequence of textual tokens as $Y = \{y_0, y_1, \ldots, y_{T-1}\}$. Specifically, we trace back the semantic relevance flow from generated tokens to raw visual inputs. For each $y_t$ at the $t$-th generation step, the relevance map $R_t$ is first initialized as an identity matrix and then iteratively updated across layers. Denote $A_t^l$ and $\nabla A_t^l$ as the multi-head attention map and the corresponding gradients in the $l$-th layer, obtained during the forward and backward passes, respectively. $R_t$ is updated as: 
\begin{equation}
    R_t = R_t + \mathbb{E}_h((A_t^l \odot \nabla A_t^l)^+)\cdot R_t ,
\label{eq:relevance_map}
\end{equation}
where $\odot$ represents Hadamard product, and $\mathbb{E}_h$ is the mean across the heads dimension. The update is performed from the $0$-th layer to the last layer. In the end, the relevance of $y_t$ to visual signals can be extracted by indexing the corresponding positions in the last row of $R_t$, that is, $R_t[-1,N_s:N_s+N_v]$. Finally, we aggregate visual relevance across all time steps $t$ by averaging, obtaining the overall visual relevance scores $R_v \in \mathbb{R}^{1\times N_v}$ with respect to the current response. A detailed analysis of Eq. \ref{eq:relevance_map} is provided in Appendix~\ref{app:0}. This well-grounded importance assessment $R_v$ can then be used to rank and select visual tokens.
% We compute the relevance map from the generated language tokens to the projected visual tokens, representing the importance distribution over visual tokens. Specifically, we trace the semantic relevance flow within the LLM. Given the input, the LLM in a MLLM generates a sequence of textual tokens $Y = \{y_i\}_{i=0}^L$ via autoregressive decoding. At the generation step $t$, we save the attention maps $A_t = \{A_t^i\}_{i=0}^l$ from all layers in LLM during the forward pass to generate $y_t$, then capture the corresponding gradients $\nabla A_t = \{\nabla A_t^i \}_{i=0}^l$ during the backward pass for $y_t$. For each generated token $y_t$,  we compute a relevance map $R_t$, which indicates the MLLM’s attention allocation during generation. The relevance map $R_t$ is first initialized as an identity matrix. In each layer, an aggregated attention map is calculated by utilizing gradients to average across heads. Then $R_t$ is iteratively updated across layers using the aggregated attention maps. The computation can be formalized as:
% \begin{equation}
%     R_t = R_t + aggregation(A_t, \nabla A_t)\cdot R_t
% \label{eq:relevance_map}
% \end{equation}
% For the generated sentence $Y$, we average the visual relevance scores extracted from each $R_t$, yielding the visual tokens importance scores $R_v \in \mathbb{R}^{1\times N_v}$ with respect to the current response. This well-grounded assessment $R_v$ is further employed to rank and select visual tokens.

\noindent\textbf{Visual Token Compression Using Relevance Scores.} The importance of visual tokens related to the instruction can be ranked according to $R_v$. We can prune the less important visual tokens down to a target count of $\hat{N_v}$, resulting in compressed token embeddings $\hat{E_v}$ as LLM input. 
% Based on the ranking obtained from generated $R_v$, we prune less important visual tokens to a target count $\hat{N_v}$, resulting in the visual representations $\hat{E_v}$. Then the compressed visual representations are fed into the LLM for response generation.

\noindent\textbf{Observation.} We visualize $R_v$ and the corresponding token pruning results for LLaVA-OneVision, Qwen2-VL, and VILA1.5 in Figure~\ref{fig:visualization}. While visualizations from different MLLMs show varying appearances due to differences in how each model processes visual input, they exhibit clear common patterns — $R_v$ for each model consistently highlights the regions corresponding to the ``three curved lines'' in the video, demonstrating the robustness of our method. Moreover, experimental results show that retaining 50\% of the original visual tokens based on $R_v$ preserves over 98\% of the performance on image benchmarks and 99\% on video benchmarks (see  Section~\ref{sec:4_2} for details). We draw the following conclusion: \textit{the explanation results faithfully capture the visual information essential for the MLLM to answer the question, and retaining only the corresponding visual tokens does not compromise model performance.} 
% We conducted token compression experiments guided by it and investigated whether $R_v$ aligns with the model’s genuine focus on visual tokens. Experimental results show that retaining 50\% of the original visual tokens preserves over 98\% of the performance on image benchmarks and 99\% on video benchmarks (detailed results can be found in Section~\ref{sec:4_2}). We alse visualize the relevance maps and token pruning guided by them for Llava-OneVision, Qwen2-VL and VILA1.5 in Figure~\ref{fig:visualization}. MLLMs differ in how their $\mathbf{VM}$ process visual inputs, leading to differences in visualization. For instance, Qwen2-VL merges visual tokens between each pair of neighboring frames, which results in the same token pruning for both frames. As shown in the left, when given an instruction to query keywords in the video, the MLLM primarily focuses on the most relevant visual regions --- textual areas within the video. The corresponding token pruning visualization on the right further demonstrates that these regions are retained. Based on the above observations, we draw the following insight: \textit{relevance map can accurately quantifies the importance of each visual token related to the instruction and is adaptable to various model architectures.}

\subsection{Explainability-based Compressor Learning}
\label{sec:3_3}
The relevance map offers valuable insights into achieving token compression at the LLM input level. However, its practical application is limited by the fact that $R_v$ is derived post-hoc -- only after the model has already generated the output. To address this limitation, we propose to approximate $R_v$ using a standalone module trained independently of the MLLM. By learning to capture attention patterns and generate relevance estimates $\tilde{R_v}$, this module enables token compression prior to LLM without modifying or retraining the MLLM. Importantly, this module prioritizes efficiency: it is lightweight, requires a small amout of training data, and can be trained quickly, making it practically applicable. The pipeline for this section is shown in the second row of Figure~\ref{fig:overview}.
% As a result, it cannot be leveraged during standard forward inference.
% Assuming that the MLLM exhibits a consistent attention pattern toward visual tokens across different inputs, we design a standalone module that is trained independently to capture this behavior and generate relevance estimates $\tilde{R_v}$ at inference time. Then $\tilde{R_v}$ 
% is employed to perform token compression. 

\noindent\textbf{Model Architecture.} As shown in Eq.~\ref{eq:relevance_map}, the relevance map is essentially obtained by aggregating attention maps, suggesting that learning a mapping from attention maps to relevance maps is promising. Interestingly yet reasonably, we find in practice that a simple convolutional network applied to the first-layer attention suffices, which guarantees the compressor’s efficiency in terms of model size, training time, and computation (implementation details and efficiency analysis can be found in Appendix~\ref{app:2} and Appendix~\ref{app:3}). Formally, let $A^0$ be the first-layer attention map. Similar to~\cite{fastv_eccv,smallforlarge_arxiv}, we focus specifically on the attention scores that visual tokens receive from textual instruction tokens. Accordingly, we extract the submap $A^0_{u \to v} \in \mathbb{R}^{N_u \times N_v}$ by indexing the corresponding positions. We then average the $N_u$ scores for each visual token to obtain a compact representation, resulting in $A^0_v \in \mathbb{R}^{1 \times N_v}$.\footnote{We omit the head dimension for notational simplicity.} This averaged attention vector $A^0_v$ is subsequently fed into a 1D convolutional model $f_\theta$ to predict visual relevance:
\begin{equation}
    \tilde{R_v} = f_\theta(A^0_v) .
\end{equation}
Note that a softmax operation is applied at the end of $f_\theta$, making $\tilde{R_v}$ a probability distribution. In addition, a separate instance of $f_\theta$ is used for each MLLM, because it is trained to approximate the explainability patterns specific to that particular MLLM.

\noindent\textbf{Training Objectives.} We process $R_v$ into the training label $R_v^*$ by masking the bottom 50\% values~\cite{GuWSYXCZ21} and normalizing the remainder into a probability distribution. Instead of softmax—which yields near-uniform values due to the closeness of raw scores—we normalize by dividing each score by the total, thereby preserving relative differences. Finally, given $R_v^*$ and $\tilde{R_v}$, the Kullback–Leibler (KL) divergence is used to measure the difference, defining the loss function:
\begin{equation}
    \mathcal{L}_{KL} = \mathbf{KL}(R_v^* || \tilde{R_v}).
\end{equation}

\noindent\textbf{Oberservation.} The learned $f_\theta$ can be seamlessly integrated into the MLLM inference pipeline to generate $\tilde{R_v}$, which can guide the token compression. As shown in Figure~\ref{fig:overview}, a visualization of $R_v$ and $\tilde{R_v}$ is given in the first and second rows, along with their corresponding pruning results, respectively. One can see that $\tilde{R_v}$ closely resembles $R_v$. Important visual regions related to the question are highlighted in both maps. This observation provides evidence that the lightweight model $f_\theta$ can indeed be efficiently and effectively trained to approximate $R_v$, allowing lossless token compression at the LLM input stage. Quantitative experimental results further support this conclusion (see Section~\ref{sec:4_3} for details).

% significantly reducing the computational cost.

% The learned compressor can be directly plugged into the inference pipeline of the MLLM to perform token compression. We apply the compressor at the input stage for token compression, significantly reducing the computational cost. Extensive experiments further demonstrate that this approach not only reduces cost but also maintains strong performance (detailed results can be found in Section~\ref{sec:4_3}) In addition, as illustrated in Figure~\ref{fig:overview}, the first row presents the relevance map and corresponding token pruning results derived from explainability method. The second row shows the visualization of the relevance estimates and pruning result produced by our learned compressor. It can be observed that the attention regions generated by our compressor closely resemble those produced by the explainability method, and that the same visual areas are largely preserved in the token pruning results. This indicates that a lightweight model can effortlessly capture the distribution of the relevance map and effectively contribute to token compression. Based on the above observations, we draw the following insight: \textit{with proper selection, token compression is feasible at the LLM input stage with negligible performance loss.}
\vspace{2em}

%% file: chapters/4.experiments.tex
\vspace{-1cm}
\section{Experiments}
\subsection{Experimental Setup}
\label{sec:4_1}

\noindent\textbf{Models.} Experiments are conducted on three MLLMs with different architectures for extensive validation, \emph{i.e.}, LLaVA-OneVision-7B~\cite{llavaov_arxiv}, Qwen2-VL-7B~\cite{qwen2vl_arxiv} and VILA1.5-8B~\cite{vila_arxiv}. These models exemplify recent advances in handling high-resolution and long visual inputs. LLaVA-OneVision and Qwen2-VL support arbitrary resolution/length, with Qwen2-VL further introducing dynamic resolution and token aggregation for compact visual representations. VILA1.5 applies spatial token compression when processing images or video frames. They thus provide a strong basis for evaluating our task-related compression, which reduces instruction-related redundancy beyond their built-in instruction-agnostic compression.

% We demonstrate the effectiveness and generalization of our method by extensive experiment on MLLMs with different architectures and input config. In detail, We apply our method to three representative MLLMs: LLaVA-OneVision-7B, Qwen2-VL-7B and VILA1.5-8B. LLaVA-OneVision processes high-resolution images by dividing them into multiple crops, resulting in a large number of visual tokens. For videos, each frame is resized to a base resolution and mapped into 196 visual tokens, with a newline token appended after the final frame to mark the sequence boundary. Qwen2-VL dynamically converts inputs of arbitrary resolution or length into a variable number of visual tokens, which are subsequently compressed in both spatial and temporal dimensions. VILA maps both images and video frames into a fixed set of 196 visual tokens as the input of LLM.

\noindent\textbf{Evaluation Tasks.} We thoroughly evaluate our method on 13 widely used image and video benchmarks. For image tasks, MME~\cite{mme} (all-round capability), MMStar~\cite{mmstar} (data contamination), MMVet~\cite{mmvet} (subjective evaluation), SEED-Bench~\cite{seedbench} (all-round capability), POPE~\cite{pope} (hallucination evaluation), TextVQA~\cite{TextVQA} (OCR reasoning), and MMBench~\cite{MMBench} (all-round capability) are included, covering various aspects of MLLM performance. For video evaluation, we select Video-MME(wo sub.)~\cite{videomme}, MVBench~\cite{mvbench}, MMBench-Video~\cite{mmbenchvideo}, NExT-QA~\cite{nextqa}, and ActivityNetQA~\cite{activityqa}, providing comprehensive coverage of video understanding abilities across different tasks and video durations. For comparison, we take existing SOTA task-related token compression methods such as FastV~\cite{fastv_eccv}, PyramidDrop~\cite{pdrop_arxiv}, Dart~\cite{Dart} as the primary baselines, which perform compression in the LLM intermediate layers on visual tokens fused with textual information in the shallow layers. 

\noindent\textbf{Implementation Details.} \emph{Training $f_\theta$}. $f_\theta$ is implemented as a five-layer convolutional network, with each layer employing a 1D depthwise separable convolution~\cite{Xception}. The training data of $f_\theta$ is collected from general-domain high-quality datasets: a subset of LLaVA-Video~\cite{llava-video} for videos and a subset of Infinity-MM~\cite{Infinity-MM} for images. To ensure high diversity, we adopt a sampling strategy that covers a wide range of task types and video duration. Implementation details including the sampling strategy of training data, the generation of $R_v$, and the training procedure of $f_\theta$ are provided in Appendix~\ref{app:2}.

\emph{Inference}. We conducted all experiments on A100 GPUs (80GB) and used VLMEvalKit~\cite{vlmevalkit_mm} for benchmarking. Details of the inference settings can be found in Appendix~\ref{app:2}. Following prior works~\cite{fastv_eccv, fitandprune}, we report FLOPs as the primary metric for evaluating inference efficiency. For a fair comparison, we configure baselines for comparable FLOPs (e.g., pruning at the 2nd layer for FastV). Our method achieves a significant reduction in inference cost with only negligible additional computation. We provide a comprehensive analysis of FLOPs of our method, please refer to Appendix~\ref{app:3}.

\subsection{Effectiveness of Compression with Explainability}
\label{sec:4_2}

\begin{table*}[t]
    \caption{\textbf{The relevance $R_v$ effectively guides token compression under different retention ratios.} Avg. means the average of performance preservation ratios across all image/video benmarks.}
    \vspace{8pt}
    \renewcommand{\arraystretch}{1.4}
    \setlength{\tabcolsep}{3pt}
    \centering
    \resizebox{1\textwidth}{!}{
    \begin{tabular}{l|c|
                >{\centering\arraybackslash}p{2cm}
                >{\centering\arraybackslash}p{1.5cm}
                >{\centering\arraybackslash}p{1.5cm}|c|
                >{\centering\arraybackslash}p{2cm}
                >{\centering\arraybackslash}p{2cm}
                >{\centering\arraybackslash}p{2cm}|c}
        \toprule
        \multirow{2}{*}{\textbf{Method}} & \multirow{2}{*}{\parbox[c]{1.5cm}{\centering \textbf{Retention Ratio}}} & \multicolumn{3}{c|}{\textbf{Image Benchmark}} & \multirow{2}{*}{\textbf{Avg.(\%)}} & \multicolumn{3}{c|}{\textbf{Video Benchmark}} & \multirow{2}{*}{\textbf{Avg.(\%)}}\\
        \cline{3-5} \cline{7-9}
        & & \textbf{MME} & \textbf{MMStar} & \textbf{MMVet} && \textbf{Video-MME} & \textbf{MVBench} & \textbf{MMBench-V} & \\
        \hline
        \cellcolor{gray!20}LLaVA-OneVision & \cellcolor{gray!20}100\% & \cellcolor{gray!20}1997.7 & \cellcolor{gray!20}60.5 & \cellcolor{gray!20}48.7 & \cellcolor{gray!20}100 & \cellcolor{gray!20}53.6 & \cellcolor{gray!20}41.2 & \cellcolor{gray!20}0.41 & \cellcolor{gray!20}100\\

        \hline
        LLaVA-OneVision & 50\% & 1974.2 & 59.7 & 47.2 & \textbf{98.1} & 54.3 & 41.1 & 0.40 & \textbf{99.5} \\
        w/GAE-Based Compressor & 25\% & 1977.3 & 59.3 & 47.0 & \textbf{97.8} & 53.8 & 40.9 & 0.40 & \textbf{99.1} \\
        
        \hline
        \cellcolor{gray!20}Qwen2-VL & \cellcolor{gray!20}100\% & \cellcolor{gray!20}2295.1 & \cellcolor{gray!20}60.4 & \cellcolor{gray!20}54.0 &  \cellcolor{gray!20}100 & \cellcolor{gray!20}50.4 & \cellcolor{gray!20}51.0 & \cellcolor{gray!20}1.23 & \cellcolor{gray!20}100\\

        \hline
        Qwen2-VL & 50\% & 2297.1 & 60.3 & 53.2 & \textbf{99.5} & 51.0 & 50.7 & 1.19 & \textbf{99.1} \\
        w/GAE-Based Compressor & 25\% & 2299.1 & 58.7 & 51.7 & \textbf{97.7} & 50.3 & 49.7 & 1.17 & \textbf{97.5} \\
        
        \hline
        \cellcolor{gray!20}VILA1.5 & \cellcolor{gray!20}100\% & \cellcolor{gray!20}1700.3 & \cellcolor{gray!20}38.7 & \cellcolor{gray!20}39.3 & \cellcolor{gray!20}100 & \cellcolor{gray!20}47.3 & \cellcolor{gray!20}34.0 & \cellcolor{gray!20}1.29 & \cellcolor{gray!20}100\\

        \hline
        VILA1.5 & 50\% & 1740.5 & 37.2 & 38.0 & \textbf{98.4} & 47.9 & 34.2 & 1.26 & \textbf{99.8} \\
        w/GAE-Based Compressor & 25\% & 1722.1 & 35.7 & 35.6 & \textbf{94.7} & 47.1 & 35.1 & 1.28 & \textbf{100.7} \\
        \bottomrule
    \end{tabular}
    }
    \label{tab:gae}
\end{table*}

\begin{table*}
    \caption{\textbf{Compare explainability-based compressor on image benchmarks.} Values marked with * in Retention Ratio denote the average retention ratio across LLM layers due to multi-stage compression in PDrop.}
    \vspace{8pt}
    \renewcommand{\arraystretch}{1.3}
    \setlength{\tabcolsep}{15pt}
    \centering
    \resizebox{1\textwidth}{!}{
    \begin{tabular}{l|c|c|ccccccc|c}
        \toprule
        \multirow{2}{*}{\makebox[0.005\textwidth][l]{\textbf{Method}}} &
        \makebox[0.05\textwidth][c]{\textbf{Retention}} &
        \multirow{2}{*}{\makebox[0.05\textwidth][c]{\textbf{FLOPs}}} &
        \multirow{2}{*}{\makebox[0.005\textwidth][c]{\textbf{MME}}} &
        \multirow{2}{*}{\makebox[0.005\textwidth][c]{\textbf{MMStar}}} &
        \multirow{2}{*}{\makebox[0.005\textwidth][c]{\textbf{MMVet}}} &
        \multirow{2}{*}{\makebox[0.005\textwidth][c]{\textbf{SEED}}} &
        \multirow{2}{*}{\makebox[0.005\textwidth][c]{\textbf{POPE}}} &
        \multirow{2}{*}{\makebox[0.005\textwidth][c]{\textbf{TextVQA}}} &
        \multirow{2}{*}{\makebox[0.005\textwidth][c]{\textbf{MMBench}}} &
        \multirow{2}{*}{\makebox[0.0005\textwidth][c]{\textbf{Avg.(\%)}}} \\

        &\makebox[0.05\textwidth][c]{\textbf{Ratio}}& & & & & & & & &\\
        \hline
        \cellcolor{gray!20}LLaVA-OV & \cellcolor{gray!20}100\% & \cellcolor{gray!20}1.00$\times$& \cellcolor{gray!20}1997.7 & \cellcolor{gray!20}60.5 & \cellcolor{gray!20}48.7 & \cellcolor{gray!20}76.7 & \cellcolor{gray!20}87.4& \cellcolor{gray!20}69.3 & \cellcolor{gray!20}80.6 &\cellcolor{gray!20}100 \\
        
        LLaVA-OV w/ FastV & 50\% & 0.51$\times$ & 679.2 & 42.7 & 28.8 & 60.1 & 10.8 & 54.4 & 47.7 & 56.0\\
        LLaVA-OV w/ Pdrop & 51\%* & 0.51$\times$ & 1974.7 & \underline{55.4} & 41.7 & \underline{74.8} & \textbf{87.0} & 64.9 & \underline{79.9} & 95.1\\
        LLaVA-OV w/ Dart & 50\% & 0.51$\times$ & \underline{1977.5} & 55.2 & \underline{42.3} & 74.3 & 85.8 & \underline{65.4} & 79.2 & 95.0\\
        \textbf{LLaVA-OV w/ Ours} & 50\% & \textbf{0.48}\boldsymbol{$\times$} & \textbf{1980.8} & \textbf{57.5} & \textbf{46.2} & \textbf{75.3} & \underline{86.2} & \textbf{66.8} & \textbf{80.1} &\textbf{97.4}\\

        \hline
        LLaVA-OV w/ FastV & 25\% & 0.27$\times$ & 527.7 & 41.5 & 20.6 & 56.1 & 10.6 & 38.4 & 42.8 & 47.3\\
        LLaVA-OV w/ PDrop & 25\%* & 0.25$\times$ & 1888.3 & \underline{50.1} & 34.7 & \underline{70.4} & 79.6 & 54.8 & 75.7 & 86.3\\
        LLaVA-OV w/ Dart & 25\% & 0.27$\times$ & \underline{1905.0} & 48.7 & \underline{36.7} & 69.5 & \underline{80.9} & \underline{57.8} & \underline{76.2} & 87.5\\
        \textbf{LLaVA-OV w/ Ours} & 25\% & \textbf{0.24}\boldsymbol{$\times$} & \textbf{1965.9} & \textbf{52.1} & \textbf{41.8} & \textbf{72.7} & \textbf{81.3} & \textbf{62.3} & \textbf{78.0} &\textbf{92.1}\\
        
        \hline
        \cellcolor{gray!20}Qwen2-VL & \cellcolor{gray!20}100\% & \cellcolor{gray!20}1.00$\times$ & \cellcolor{gray!20}2295.1 & \cellcolor{gray!20}60.4 & \cellcolor{gray!20}54.0 & \cellcolor{gray!20}75.8 & \cellcolor{gray!20}87.5 & \cellcolor{gray!20}84.1
        & \cellcolor{gray!20}81.0 &\cellcolor{gray!20}100 \\
        
        Qwen2-VL w/ FastV & 50\% & 0.51$\times$ & 1489.3 & 41.4 & 34.4 & 56.0 & 83.0 & 66.7 & 42.1 & 71.0\\
        Qwen2-VL w/ PDrop & 51\%* & 0.51$\times$ & 2288.1 & 55.4 & 46.3 & \underline{73.0} & 86.3 & 80.9 & 79.7 & 95.2\\ 
        Qwen2-VL w/ Dart & 50\% & 0.51$\times$ & \textbf{2290.0} & \underline{55.5} & \underline{49.4} & 72.4 & \textbf{86.6} & \textbf{82.7} & \underline{80.1} & 96.4 \\
        \textbf{Qwen2-VL w/ Ours} & 50\% & \textbf{0.49}\boldsymbol{$\times$} & \underline{2288.3} & \textbf{55.9} & \textbf{51.9} & \textbf{73.2} & \underline{86.4} & \underline{82.6} & \textbf{80.9} &\textbf{97.4} \\
        \hline
        Qwen2-VL w/ FastV & 25\% & 0.27$\times$ & 1415.3 & 37.6 & 31.4 & 51.4 & 77.6 & 63.2 & 39.1 & 66.0\\
        Qwen2-VL w/ PDrop & 25\%* & 0.25$\times$ & \underline{2216.3} & 51.1 & 42.3 & 67.4 &83.2 & 75.4 & 77.0 & 89.7\\
        Qwen2-VL w/ Dart & 25\% & 0.27$\times$ & 2184.5 & \underline{51.3} & \underline{45.6} & \underline{68.2} & \underline{84.3} & \underline{76.7} & \underline{77.5} & 91.1\\
        \textbf{Qwen2-VL w/ Ours} & 25\% & \textbf{0.24}\boldsymbol{$\times$} & \textbf{2280.9} & \textbf{51.8} & \textbf{47.3} & \textbf{67.9} & \textbf{84.8} & \textbf{80.0} & \textbf{77.7} & \textbf{92.9}\\
        
        \bottomrule
    \end{tabular}
    }
    \label{tab:image_base}
\end{table*}
% We compare our compressor, FastV and vanilla model under different retention ratios across four image benchmarks. 

We conduct experiments to verify whether the explanation results can guide token compression, \emph{i.e.}, compressing $E_v$ to $\hat{E_v}$ according to $R_v$ and then feeding $\hat{E_v}$ into $\mathbf{LM}$ to generate a response. To assess effectiveness and generalization, we apply the method to three state-of-the-art MLLMs and test them on three image and three video benchmarks. 

Table~\ref{tab:gae} reports the results under retention ratios of 50\% and 25\%. The strong performance across multiple models and benchmarks demonstrates the effectiveness and broad applicability of such an explainability-based token compressor. For Qwen2-VL, reducing visual tokens by 50\% still preserves over 99\% of the original performance on both image and video tasks. LLaVA-OneVision retains 99.1\% of its video performance even with only 25\% of tokens. VILA reduces visual tokens to 98 per image or frame at 50\% retention, while maintaining 98\% of the original image performance and nearly unchanged video performance. These observations indicate that token compression based on relevance $R_v$ effectively preserves the visual tokens essential for MLLMs to answer the question.

\subsection{Effectiveness of Explainability-based Compressor Learning}
\label{sec:4_3}
The performance of the $\tilde{R_v}$-guided token compressor is evaluated in this section. $\tilde{R_v}$ is generated by the learned $f_\theta$, and the token pruning is performed accordingly before the LLM computation. Five image and six video benchmarks are included for evaluation. 
% For a comprehensive evaluation, we compares our compressor with FastV on image and video benchmarks respectively. FastV prunes tokens at shallow layer (layer = 3 in our setup) of LLM based on attention scores from query tokens. In contrast, our compressor prunes irrelevant tokens prior to feeding them into the LLM.

% \noindent\textbf{Performance under Constrained Visual Input Configuration.} 
\noindent\textbf{Performance Comparison.} 
Table~\ref{tab:image_base} presents the results of LLaVA-OneVision and Qwen2-VL under different token compression retention ratios on image benchmarks. We exclude VILA here because it uses a fixed and relatively small number of image tokens, making compression less meaningful. As shown in the table, at a retention rate of 50\%, our compressor demonstrates overall superiority over the baselines at comparable FLOPs, achieving average improvements of 2.6\% and 1.2\% across all benchmarks for LLaVA-OneVision and Qwen2-VL, respectively. When the retention rate is further reduced to 25\%, the performance gains increase to 4.7\% and 1.6\%, highlighting the enhanced robustness of our method under higher compression rates. We conducted additional experiments on more image benchmarks; please refer to the Appendix~\ref{app:4}.

% Table~\ref{tab:image_base} presents the result of LLaVA-OneVision and Qwen2-VL under different retention ratios of token compression on an image benchmark. We exclude VILA from experiments on image tasks due to its fixed and relatively small image token count (196 token for a single image), which renders compression less meaningful. At a retention rate of 50\%, our compressor outperforms FastV by 0.5\% and 0.1\% on LLaVA-OneVision and Qwen2-VL, respectively. When the retention rate is further reduced to 25\%, the performance margins increase to 3.6\% and 0.8\%, demonstrating enhanced robustness under higher compression rates and highlighting an optimal trade-off between efficiency and accuracy.
% 补充对某些数据集如MMStar单独分析？

In Table~\ref{tab:video_base}, we evaluate the compression performance of LLaVA-OneVision, Qwen2-VL, and VILA on video benchmarks. We make several observations. First, our compressor consistently outperforms baselines at comparable FLOPs, regardless of the model and retention ratio. Both LLaVA-OneVision and VILA are able to maintain 100\% performance when 50\% of the visual tokens are pruned. Second, VILA exhibits the smallest performance drop, while Qwen2-VL shows the largest, likely due to its attention patterns being harder to capture. Importantly, our task-related compression can still further reduce token redundancy on both Qwen2-VL and VILA, demonstrating that it complements the task-agnostic compression already presented in these models. Finally, comparing the results in Tables~\ref{tab:gae}, \ref{tab:image_base}, and \ref{tab:video_base}, the performance degradation from the $R_v$-guided compressor to the $\tilde{R_v}$-guided compressor is more pronounced in image tasks. This is likely also due to the greater redundancy in videos, which reduces the learning difficulty.

% This suggests that the visual representations of Qwen2-VL are the most difficult to compress, probably because it has already obtained compact and highly aggregated representations through a series of operations. 

% In Table~\ref{tab:video_base}, we evaluate the performance of LLaVA-OneVision, Qwen2-VL, and VILA on video benchmarks. Compared to images, videos typically exhibit lower information density with respect to the instruction, which motivates the inclusion of a lower retention ratio of 10\% in video experiments. At relatively high token retention ratios (i.e., 50\%), our compressor achieves superior average performance over the vanilla baseline  on both LLaVA-OneVision and VILA. This advantage is particularly pronounced in VILA, where the compressor dominates nearly all benchmarks, indicating that a substantial portion of instruction-irrelevant visual information in videos may negatively impact response quality. When the token retention ratio is decreased to 25\% and 10\%, our compressor consistently outperforms FastV on average across video benchmarks. This suggests that the explainability-driven compressor more effectively preserves critical visual information, mitigating performance degradation under aggressive compression. Across all token retention ratios, Qwen-2VL demonstrates inferior performance compared to LLaVA-OneVision and VILA. We attribute this performance gap to Qwen-2VL's highly fused visual token representations.

\begin{table*}
    \caption{\textbf{Compare explainability-based compressor on video benchmarks.} As videos generally exhibit greater visual redundancy, we also evaluate a lower retention ratio 10\% to further assess compression robustness, with detailed results reported in the Appendix~\ref{app:4}. }
    \vspace{8pt}
    \renewcommand{\arraystretch}{1.3}
    \setlength{\tabcolsep}{1pt}
    \centering
    \resizebox{1\textwidth}{!}{
    \begin{tabular}{l|c|c|cccccc|c}
        \toprule
        % 可以加一行type
        \multirow{2}{*}{\makebox[0.2\textwidth][l]{\textbf{Method}}}&
        \multirow{2}{*}{\parbox[c]{1.5cm}{\centering \textbf{Retention Ratio}}}& 
        \multirow{2}{*}{\makebox[0.1\textwidth][c]{\textbf{FLOPs}}} &
        \multirow{2}{*}{\makebox[0.15\textwidth][c]{\textbf{Video-MME}}} &
        \multirow{2}{*}{\makebox[0.15\textwidth][c]{\textbf{MVBench}}} &
        \multirow{2}{*}{\parbox[c]{1.5cm}{\centering \textbf{MMBench-Video}}} &
        \multicolumn{2}{c}{\makebox[0.2\textwidth][c]{\textbf{Next-QA}}} &
        \multirow{2}{*}{\makebox[0.15\textwidth][c]{\textbf{Activity-QA}}} &
        \multirow{2}{*}{\makebox[0.1\textwidth][c]{\textbf{Avg.(\%)}}}\\
        \cline{7-8} 
        & & & & & & \textbf{multi-choice} & \textbf{open-ended} & & \\
        \hline
        \cellcolor{gray!20}LLaVA-OV & \cellcolor{gray!20}100\% & \cellcolor{gray!20}1.00$\times$ & \cellcolor{gray!20}53.6 & \cellcolor{gray!20}41.2 & \cellcolor{gray!20}0.41 & \cellcolor{gray!20}79.2 & \cellcolor{gray!20}49.0 & \cellcolor{gray!20}56.9 & \cellcolor{gray!20}100\\
        
        LLaVA-OV w/ FastV & 50\% & 0.48$\times$ & 42.3 & 25.0 & 0.35 & 66.5 & 36.0 &48.5 & 78.0\\
        LLaVA-OV w/ PDrop & 50\%* & 0.47$\times$ & 52.7 & \underline{40.2} & 0.36 & \underline{78.4} & 48.2 & 56.0 & 96.6\\
        LLaVA-OV w/ Dart & 50\% & 0.48$\times$ & \underline{53.2} & 40.0 & \underline{0.36} & 78.0 & \underline{49.0} & \underline{56.2} & 96.9 \\
        \textbf{LLaVA-OV w/ Ours} & 50\% & \textbf{0.46}\boldsymbol{$\times$} & \textbf{53.4} & \textbf{40.5} & \textbf{0.43} & \textbf{78.6} & \textbf{49.7} & \textbf{56.5} & \textbf{100.4} \\

        \hline
        LLaVA-OV w/ FastV & 25\% & 0.25$\times$ & 39.6 & 23.6 & 0.30 & 64.2 & 33.6 & 44.5 & 72.0 \\
        LLaVA-OV w/ PDrop & 25\%* & 0.24$\times$ & 50.8 & 38.2 & \underline{0.35} & 76.3 & \underline{48.2} & 53.5 & 93.6 \\
        LLaVA-OV w/ Dart & 25\% & 0.25$\times$ & \textbf{51.5} & \underline{38.7 }& 0.33 & \underline{76.6} & 47.0 & \textbf{55.1} & 93.3 \\
        \textbf{LLaVA-OV w/ Ours} & 25\% & \textbf{0.22}\boldsymbol{$\times$} & \underline{51.3} & \textbf{39.0} & \textbf{0.42} & \textbf{77.0} & \textbf{49.0} & \underline{54.5} & \textbf{97.3} \\
        
        \hline
        \cellcolor{gray!20}Qwen2-VL & \cellcolor{gray!20}100\% & \cellcolor{gray!20}1.00$\times$ & \cellcolor{gray!20}50.4 & \cellcolor{gray!20}51.0 & \cellcolor{gray!20}1.23 & \cellcolor{gray!20}76.8 & \cellcolor{gray!20}45.5 & \cellcolor{gray!20}53.6 & \cellcolor{gray!20}100\\
        
        Qwen2-VL w/ FastV & 50\% & 0.48$\times$ & 32.4 & 36.3 & 0.52 & 43.9 & 28.3 & 38.2 & 61.4 \\
        Qwen2-VL w/ PDrop & 50\%* & 0.47$\times$ & 48.9 & 49.6 & 1.14 & 75.2 & \underline{45.4} & 50.8 & 96.6 \\
        Qwen2-VL w/ Dart & 50\% & 0.48$\times$ & \underline{49.6} & \underline{49.4} & \underline{1.17} & \underline{76.4} & 44.8 & \underline{52.0} & 97.6 \\
        \textbf{Qwen2-VL w/ Ours} & 50\% & \textbf{0.46}\boldsymbol{$\times$} & \textbf{50.0} & \textbf{49.8} & \textbf{1.18} & \textbf{75.6} & \textbf{45.9} & \textbf{52.4} & \textbf{98.3} \\

        \hline
        Qwen2-VL w/ FastV & 25\% & 0.25$\times$ & 31.2 & 36.1 & 0.48 & 42.0 & 26.8 & 35.1 & 58.5 \\
        Qwen2-VL w/ PDrop & 25\%* & 0.24$\times$ & 47.3 & 46.2 & \underline{1.11} & 73.9 & 44.1 & 47.8 & 92.8 \\
        Qwen2-VL w/ Dart & 25\% & 0.25$\times$ & \underline{47.4} & \underline{47.1} & 1.10 & \underline{74.1} & \textbf{44.5} & \underline{49.2} & 93.7 \\
        \textbf{Qwen2-VL w/ Ours} & 25\% & \textbf{0.22}\boldsymbol{$\times$} & \textbf{48.1} & \textbf{46.7} & \textbf{1.11} & \textbf{74.2} & \underline{44.3} & \textbf{50.5} & \textbf{94.2} \\

        \hline
        \cellcolor{gray!20}VILA & \cellcolor{gray!20}100\% & \cellcolor{gray!20}1.00$\times$ & \cellcolor{gray!20}47.3 & \cellcolor{gray!20}34.0 & \cellcolor{gray!20}1.29 & \cellcolor{gray!20}69.9 & \cellcolor{gray!20}46.2 & \cellcolor{gray!20}55.6 & \cellcolor{gray!20}100\\
        
        VILA w/ FastV & 50\% & 0.49$\times$ & 42.2 & 20.7 & 0.98 & 62.9 & 36.8 &47.1& 80.1 \\
        VILA w/ PDrop & 50\%* & 0.49$\times$ & \underline{47.3} & \underline{35.0} & 1.22 & \underline{69.4} & 45.8 & 55.1 & 99.2 \\
        VILA w/ Dart & 50\% & 0.49$\times$ & 46.1 & 34.7 & \underline{1.25} & 69.2 & \underline{46.5} &\underline{55.2}& 99.2 \\
        \textbf{VILA w/ Ours} & 50\% & \textbf{0.47}\boldsymbol{$\times$} & \textbf{47.6} & \textbf{35.2} & \textbf{1.25} & \textbf{70.3} & \textbf{46.4} & \textbf{55.4} & \textbf{100.3}\\

        \hline
        VILA w/ FastV & 25\% & 0.26$\times$ & 41.4 & 20.5 & 0.97 & 61.5 & 36.4 &46.8& 79.0 \\
        VILA w/ PDrop & 25\%* & 0.26$\times$ & 45.2 & 33.6 & \underline{1.24} & \underline{68.2} & 45.1 &\underline{54.8}& 97.4 \\
        VILA w/ Dart & 25\% & 0.26$\times$ & \underline{45.3} & \underline{34.6} & 1.23 & 68.1 & \underline{45.6} &54.2& 97.7 \\
        \textbf{VILA w/ Ours} & 25\% & \textbf{0.23}\boldsymbol{$\times$} & \textbf{45.5} & \textbf{35.6} & \textbf{1.22} & \textbf{69.4} & \textbf{46.4} & \textbf{54.8} & \textbf{99.0} \\
        
        \bottomrule
    \end{tabular}}
    \label{tab:video_base}
\end{table*}
% We extend our evaluation to a lower retention ratio of 10\% on video tasks and find that our compressor consistently outperforms FastV on average across video benchmarks.

\begin{figure*}[t]
  \centering
  \includegraphics[width=\textwidth]{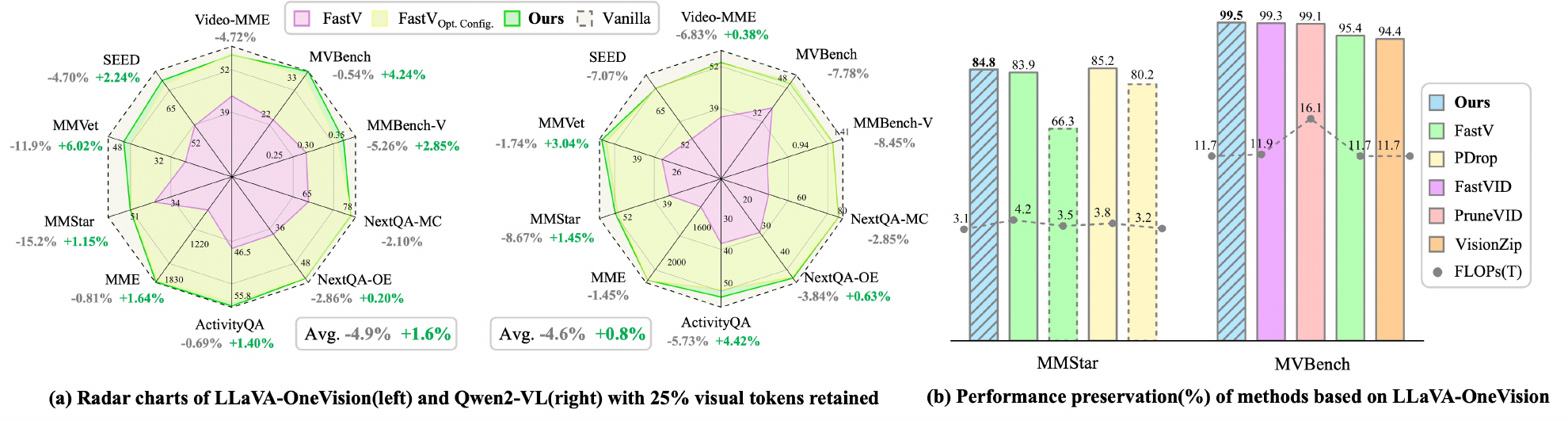}
  \caption{\textbf{Comparison results on larger images and longer videos.} Performance preservation ratio measures the performance retained relative to Vanilla. \textcolor{mygray}{Gray text} denotes the gap from Vanilla and \textcolor{mygreen}{green text} highlights improvements over the FastV\textsubscript{Opt.Config.}.}
  \label{fig:all}
\end{figure*}

\begin{table*}
    \vspace{-0.8cm}    
    \caption{\textbf{Efficiency analysis based on Qwen2-VL on MMStar.} We evaluate the inference costs in terms of total inference time, prefilling time, FLOPs, and KV cache memory. KV cache memory is computed with consideration of the Grouped Query Attention (GQA) used in practical inference.}
    \vspace{8pt}
    \renewcommand{\arraystretch}{1.2}
    \setlength{\tabcolsep}{1pt}
    \centering

    \resizebox{1\textwidth}{!}{
    \begin{tabular}{l|c|c|ccccc|c}
        \toprule
        \multirow{2}{*}{\makebox[0.25\textwidth][l]{\textbf{Method}}}&
        \multirow{2}{*}{\parbox[c]{1.5cm}{\centering \textbf{Retention Ratio}}}& 
        \multirow{2}{*}{\makebox[0.15\textwidth][c]{\textbf{FLOPs(\boldsymbol{$\times$})}}} &
        \multirow{2}{*}{\parbox[c]{2.5cm}{\centering \textbf{Total Inference Time}}} &
        \multirow{2}{*}{\parbox[c]{1.5cm}{\centering \textbf{Prefilling Time}}} &
        \multirow{2}{*}{\parbox[c]{2cm}{\centering \textbf{KV Cache}}} &
        \multirow{2}{*}{\parbox[c]{2cm}{\centering \textbf{Total Speedup}}} &
        \multirow{2}{*}{\parbox[c]{2cm}{\centering \textbf{Prefilling Speedup}}} &
        \multirow{2}{*}{\parbox[c]{1.5cm}{\centering \textbf{MMStar}}} \\
        &&&&&&&&\\
        \hline
        \cellcolor{gray!20}Qwen2-VL & \cellcolor{gray!20}100\% & \cellcolor{gray!20}1.00$\times$ & \cellcolor{gray!20}15min24s & \cellcolor{gray!20}6min36s & \cellcolor{gray!20}71.2MB & \cellcolor{gray!20}1.00$\times$ & \cellcolor{gray!20}1.00$\times$& \cellcolor{gray!20}61.1 \\
        
        Qwen2-VL w/ FastV & 25\% & 0.27$\times$ & 12min19s & 4min14s & 19.7MB & 1.25$\times$ & 1.56$\times$ &39.6\\
        Qwen2-VL w/ PDrop & 25\%* & 0.25$\times$ & 12min15s & 4min10s & 18.1MB & 1.26$\times$ & 1.58$\times$ & 53.1\\
        Qwen2-VL w/ Dart & 25\% & 0.30$\times$ & 12min20s & 4min16s & 21.6MB & 1.25$\times$ & 1.55$\times$ & 54.3\\
        \textbf{Qwen2-VL w/ Ours} & 25\% & \textbf{0.24}\boldsymbol{$\times$} & \textbf{12min16s} & \textbf{4min08s} & \textbf{17.8MB} & \textbf{1.26}\boldsymbol{$\times$} & \textbf{1.60}\boldsymbol{$\times$} & \textbf{55.8}\\
        \bottomrule
    \end{tabular}
    }
    \label{tab:efficiency}
\end{table*}

\noindent\textbf{Applying to Larger Images and Longer Videos.} 
We evaluate the generalization of directly applying the trained $f_\theta$ on larger images and longer videos; experimental results are shown in Figure~\ref{fig:all}, with configurations detailed in Appendix~\ref{app:2}. Sub-figure(a) show the compression performance on 4 image and 6 video benchmarks based on LLaVA-OneVision and Qwen2-VL. Our method consistently outperforms FastV and achieves higher average performance than its optimal configuration, demonstrating strong generalization to larger images and longer videos than it seen during training. Detailed comparison results on these 10 benchmarks are provided in the Appendix~\ref{app:4}.
% For example, although it is trained only on videos with 8 frames, it can be directly applied to token compression for videos with 32 frames, achieving excellent performance (see Appendix~\ref{app:2} for implementation details).

Sub-figure(b) show the comparisons on two challenging benchmarks, \emph{i.e.}, MMStar and MVBench. Several strong methods are introduced for comparison: PruneVID~\cite{prunevid_arxiv}, FastVID~\cite{fastvid_arxiv}, VisionZip~\cite{visionzip_arxiv}, and PyramidDrop~\cite{pdrop_arxiv}. Our approach achieves SOTA performance with the lowest FLOPs. Remarkably, even directly applying the trained $f_\theta$ to longer videos with more frames, it still performs favorably compared to methods specifically designed for videos (i.e., FastVID and PruneVID). Beyond superior performance, our lightweight compressor significantly improves MLLM inference efficiency with negligible additional cost. We follow Dart~\cite{Dart} and report efficiency in terms of total inference time, prefilling time, FLOPs, and KV cache memory, as shown in Table~\ref{tab:efficiency}. Our compressor achieves both the best performance and highest efficiency. The prefill-stage acceleration (1.60$\times$) and reduced KV cache footprint (71.2MB \textrightarrow 17.8MB) enable efficient processing in prefill and decode stages, keeping total inference time comparable to other methods (see Appendix~\ref{app:4} for more results).

\noindent\textbf{Analysis and Discussion. }We find it interesting that a lightweight $f_\theta$ can achieve strong performance, thus we offer further discussion here: (i) \textit{The convolutional mapper $f_\theta$ learns a structurally matched input–output mapping. }As shown in Eq.~\ref{eq:relevance_map}, the relevance map $R_t$ is essentially an iterative composition of attention maps $A_t^l$ with their task-specific gradients across layers. Therefore, using $A$ to predict $R$ is natural and intuitively well-motivated. Visualization results in Appendix~\ref{app:1} show that $f_\theta$ exhibits task-sensitive behavior, adaptively focusing on the relevant visual regions according to the instruction. (ii) \textit{Shallow-layer attention maps contain vital information. }It has been demonstrated in several works~\cite{fastv_eccv, pdrop_arxiv} that visual tokens make a greater contribution to output generation in the shallow layers compared to the deeper layers. Our experimental results are consistent with the finding. First, using the first-layer attention map as the input of $f_\theta$ already provides a strong guidance for token compression. Moreover, the attention maps from shallow layers perform well overall; for detailed ablation results, please refer to the Appendix~\ref{app:6}. (iii) \textit{We aim not to learn an identical $R_t$, but rather its relatively large values. }The loss function design in Section~\ref{sec:3_3} masked the bottom 50\% of label values, simplifying the learning task to the distribution of the top 50\%. Our goal is not to learn a precise one-to-one input–output mapping. Instead, the learning target is only to achieve identifying which regions are relevant to the instruction, thereby enabling token pruning. One can observe the visualization results in Appendix~\ref{app:1} for evidence. The pruning results ${\scriptstyle\hat{V}}$(based on ${\scriptstyle R_v}$) and the convolutional network’s learned ${\scriptstyle\hat{\tilde{V}}}$(based on ${\scriptstyle\tilde{R_v}}$) are not identical, yet it retains the same task-related regions and produces correct answers.

\begin{table*} 
    \caption{\textbf{Ablation study on explainability methods for relevance map generation.} We evaluate two strategies for aggregating multi-head attention maps—gradient-weighted summation and simple averaging—to generate relevance maps for token compression on video and image benchmarks.}
    \vspace{8pt}
    \renewcommand{\arraystretch}{1.3}
    \setlength{\tabcolsep}{3pt}
    \centering
    \resizebox{1\textwidth}{!}{
    \begin{tabular}{c|c|c|
                >{\centering\arraybackslash}p{1.5cm}
                >{\centering\arraybackslash}p{1.5cm}
                >{\centering\arraybackslash}p{1.5cm}|
                >{\centering\arraybackslash}p{2cm}
                >{\centering\arraybackslash}p{2cm}
                >{\centering\arraybackslash}p{2cm}|c}
        \toprule
        \multirow{2}{*}{\textbf{Model}} & \multirow{2}{*}{\textbf{Method}} & \multirow{2}{*}{\parbox[c]{1.5cm}{\centering \textbf{Rentation Ratio}}} & \multicolumn{3}{c|}{\textbf{Image Benchmark}} & \multicolumn{3}{c|}{\textbf{Video Benchmark}} & \multirow{2}{*}{\textbf{Avg.(\%)}}\\
        \cline{4-6} \cline{7-9}
        & & & \textbf{MME} & \textbf{MMStar} & \textbf{MMVet} & \textbf{Video-MME} & \textbf{MVBench} & \textbf{MMBench-V} & \\
        \hline
        \multirow{3}{*}{\parbox[c]{2cm}{\centering LLaVA-OneVision}} & \cellcolor{gray!20} Vanilla & \cellcolor{gray!20}100\% & \cellcolor{gray!20}1997.7 & \cellcolor{gray!20}60.5 & \cellcolor{gray!20}48.7 & \cellcolor{gray!20}53.6 & \cellcolor{gray!20}41.2 & \cellcolor{gray!20}0.41 & \cellcolor{gray!20}100\\
        \cline{2-10} 
        & Mean-weighted & \multirow{2}{*}{50\%} & 1974.5 & 58.5 & 45.9 & 53.6 & 40.8 & 0.39 & 97.3 \\
        & \textbf{Grad-weighted} & & 1974.2 & 59.7 & 47.2 & 54.3 & 41.1 & 0.40  & \textbf{98.8}\\
        \cline{2-10} 
        \hline
        \multirow{3}{*}{Qwen2-VL} & \cellcolor{gray!20} Vanilla & \cellcolor{gray!20}100\% & \cellcolor{gray!20}2295.1 & \cellcolor{gray!20}60.4 & \cellcolor{gray!20}54.0 & \cellcolor{gray!20}50.4 & \cellcolor{gray!20}51.0 & \cellcolor{gray!20}1.23 & \cellcolor{gray!20}100\\
        \cline{2-10} 
        & Mean-weighted & \multirow{2}{*}{50\%} & 2300.6 & 58.2 & 49.2 & 49.9 & 49.9 & 1.15 & 96.3 \\
        & \textbf{Grad-weighted} & & 2297.1 & 60.3  & 53.2  & 51.0 & 50.7  & 1.19 & \textbf{99.3}\\
        \cline{2-10} 
        \hline
        \multirow{3}{*}{VILA1.5} & \cellcolor{gray!20} Vanilla & \cellcolor{gray!20}100\% & \cellcolor{gray!20}1700.3 & \cellcolor{gray!20}38.7 & \cellcolor{gray!20}39.3 & \cellcolor{gray!20}47.3 & \cellcolor{gray!20}34.0 & \cellcolor{gray!20}1.29 & \cellcolor{gray!20}100\\
        \cline{2-10} 
        & Mean-weighted & \multirow{2}{*}{50\%} & 1720.8 & 38.0 & 34.2 & 48.0 & 34.1 & 1.20 & 96.9 \\
        & \textbf{Grad-weighted} & & 1740.5 & 37.2 & 38.0 & 47.9 & 34.2 & 1.26 & \textbf{99.1}\\
        \bottomrule
    \end{tabular}
}
    \label{tab:ablation}
\end{table*}

\vspace{-0.3cm}
\subsection{Ablation Study}
\label{sec:4_4}
As shown in Eq. \ref{eq:relevance_map}, the relevance map is updated by aggregating attention maps across layers, where the multiple heads in each layer are combined either via simple averaging or gradient-weighted averaging (used in our approach). Table~\ref{tab:ablation} shows that employing
gradient-weighted aggregation to generate $R_v$ for token compression performs consistently better than simple averaging across image and video benchmarks. A reasonable explanation is that differing contributions of attention heads make simple averaging prone to distorting relevance maps~\cite{multihead_acl}. We also include ablation studies of different configurations for training $f_\theta$ in Appendix~\ref{app:6}, specifically investigating (i) the effect of varying the depth of the convolutional network and (ii) the influence of selecting different attention layer as input for the compressor.

% In this part, we mainly study the influence of aggregation methods in explainability methods on the quality of generated relevance maps. The relevance map is updated ulitizing the attention map of each attention layer. Since each attention map is comprised of multiple heads, the heads can be fused by simple averaging, or by using gradient-weighted aggregation. As shown in the Table~\ref{tab:ablation}, we evaluated the guiding ability of relevance maps generated by both methods on token compression. Whether in image or video benchmarks, the gradient-weighted aggregation method consistently outperforms others in terms of compression results, demonstrating that it produces higher-quality relevance maps that more accurately capture the distribution of importance across visual tokens. This can be explained by the fact that attention heads vary in their importance and relevance, meaning a simple average across heads leads to distorted relevance maps~\cite{multihead_acl}.Therefore, we employ a gradient-weighted aggregation strategy, as simple averaging, despite its ease of implementation, often results in relevance maps of inferior quality.

%% file: chapters/5.conclusion.tex
\vspace{-0.2cm}
\section{Conclusion}
In this work, we demonstrate the feasibility of task-related visual token compression at the LLM input stage. We first demonstrate experimentally that the relevance scores derived from explainability methods well evaluate the task-related importance of visual tokens, which can be used for effective token compression. To enable efficient and practical deployment, we employ a simple convolutional network to learn a mapping from the LLM first-layer attention maps to the explainability-derived relevance scores. Using the predicted relevance scores from lightweight model, token compression can be performed prior to the LLM. Extensive experiments demonstrate the effectiveness and generalizability of our task-related token compression method. Since the relevance scores are obtained via backward computations, their generation is resource-intensive. This poses a challenge in scaling the compressor training to high-resolution images or long video sequences. Future work will explore stronger compressors and the use of relevance scores to guide token compression during training.

%% file: chapters/appendix.tex
%%%%%%%%%%%%%%%
% \documentclass{article}
% \usepackage{neurips_2025}

% \usepackage[utf8]{inputenc} % allow utf-8 input
% \usepackage[T1]{fontenc}    % use 8-bit T1 fonts
% \usepackage{hyperref}       % hyperlinks
% \usepackage{url}            % simple URL typesetting
% \usepackage{booktabs}       % professional-quality tables
% \usepackage{amsfonts}       % blackboard math symbols
% \usepackage{nicefrac}       % compact symbols for 1/2, etc.
% \usepackage{microtype}      % microtypography
% \usepackage{xcolor}         % colors

% \usepackage{bm}
% \usepackage{enumitem}
% \usepackage{amsmath}
% \usepackage{multirow}
% \usepackage{multicol}
% \usepackage{graphicx}
% \usepackage{tabularx}
% \usepackage{floatrow}

% \begin{document}
%%%%%%%%%%%%%%%
\appendix
\section*{Appendix}

\section{Details of Relevance Propagation Equation}
\label{app:0}
The relevance propagation in Eq.~\ref{eq:relevance_map} follows the Generic Attention Explainability (GAE) framework~\cite{gae_iccv}, which is a powerful method to interpret predictions for Transformer-based architectures.
We trace the contribution of each visual token to the model’s output by leveraging the self-attention modules within the GAE framework to assess token importance. Specifically, GAE first generates a relevance map $\bar{A^l}$ for each layer $l$ by integrating raw attention map and its gradients with respect to the current output $y$:
\begin{equation}
    \bar{A^l}=\mathbb{E}_h((A^l \odot \nabla A^l)^+),
\label{eq:layer_relevance_map}
\end{equation}
where $A^l$ can be obtained through a forward pass and the related gradient $\nabla A^l := \frac{\partial y_t}{\partial A^l}$ can be cached during a backward pass. GAE computes the Hadamard product of the attention map and its gradient for the intuitive reasons that: 1) it captures how much attention each token receives from other tokens (information from the attention map), 2) it identifies which tokens require more attention to effectively influence the output (information from the gradient). $(·)^+$ represents the operation of setting negative values to 0. By zeroing out negative gradient components, GAE enforces a causal direction of influence and prevents the suppression of informative positive signals by accumulated negative components~\cite{CheferGW21, Grad-SAM}. Gradient can also be viewed as weight in the head aggregation to indicate corresponding importance~\cite{multihead_acl}, thereby $\mathbb{E}_h$ performs a gradient-weighted expectation over the head dimension.

Since the attention module is followed by a residual connection, GAE accumulate the relevance map by adding each layer’s contribution $\bar{A^l}$ to the aggregated relevance map $R$:
\begin{equation}
    R = R+\bar{A^l} \cdot R,
\label{eq:accumulate_relevance_map}
\end{equation}
The overall relevance map $R$ is initialized as the identity matrix with the intuition that each input token’s relevance score is identical in the beginning. Then the relevance propagation updates the $R$ from the 0-th layer to the last layer in the Transformer.

The traditional GAE method lacks a mechanism for handling the token-by-token autoregressive outputs required by MLLMs. Therefore, we adopt the stepwise relevance computation from R-GAE~\cite{deco_arxiv}, a GAE-derived method adapted to MLLMs. For a generated sequence of textual tokens $Y = \{y_0, y_1, \ldots, y_{T-1}\}$, we obtain the overall relevance by averaging the GAE relevance maps computed using Eq.~\ref{eq:accumulate_relevance_map} at each decoding step $t$.

\section{More Visualization Results}
\label{app:1}
\subsection{Visualization Results Across Different MLLMs}
\begin{figure*}[ht]
  \centering
  \includegraphics[width=\textwidth]{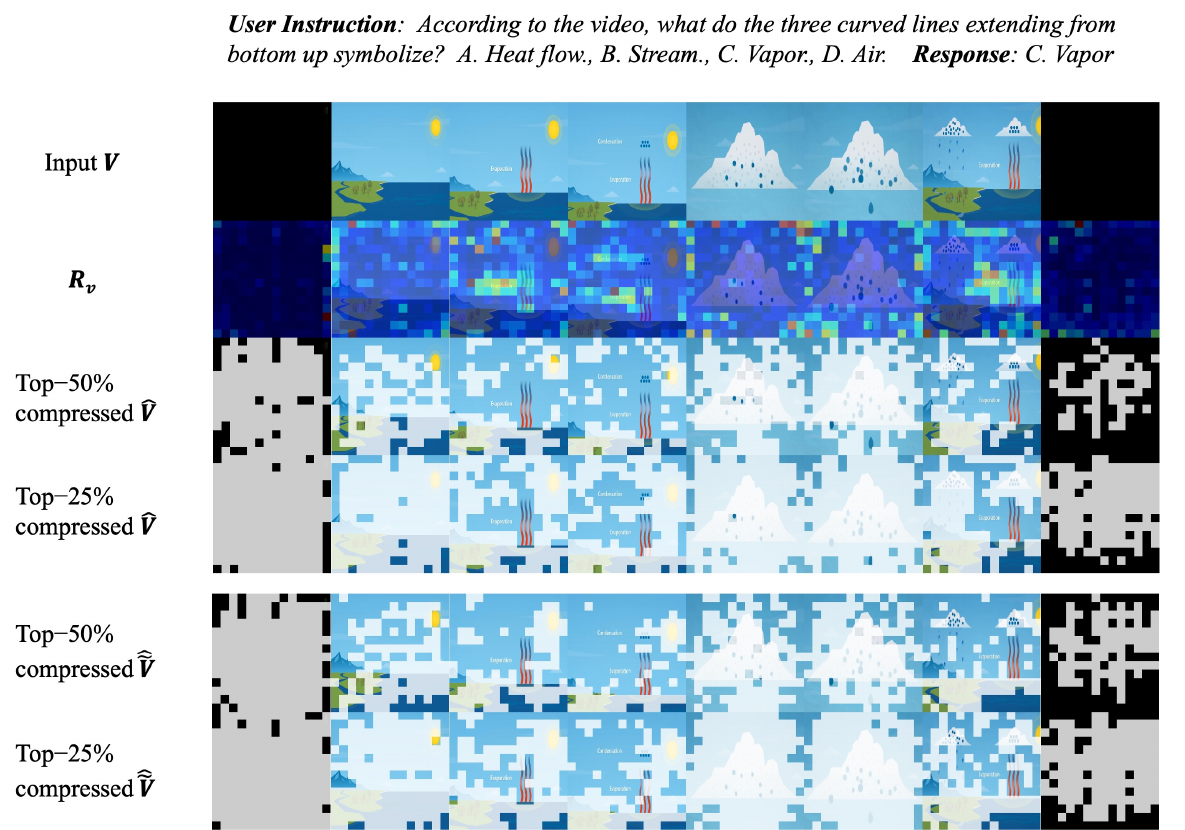}
  \caption{\textbf{Video Input Visualizations for LLaVA-OneVision}. }
  \label{fig:llavaov_video}
\end{figure*}

\begin{figure*}[ht]
  \centering
  \includegraphics[width=\textwidth]{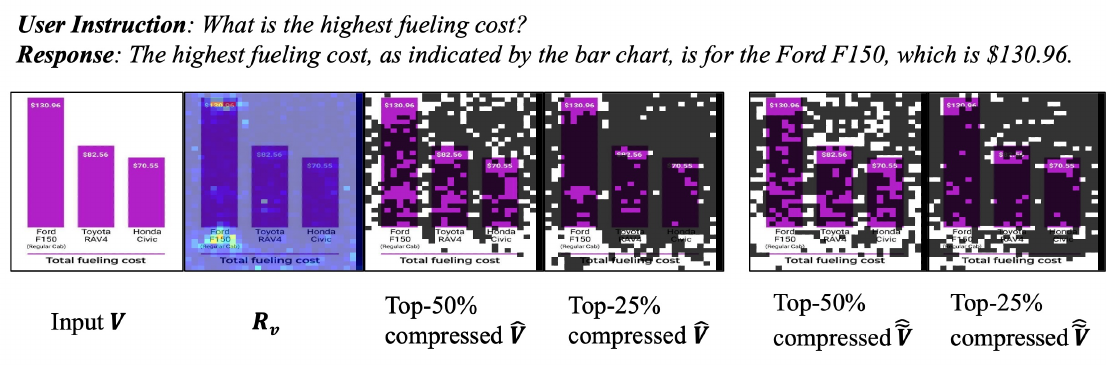}
  \caption{\textbf{Image Input Visualizations for LLaVA-OneVision}. }
  \label{fig:llavaov_image}
\end{figure*}

\begin{figure*}[ht]
  \centering
  \includegraphics[width=\textwidth]{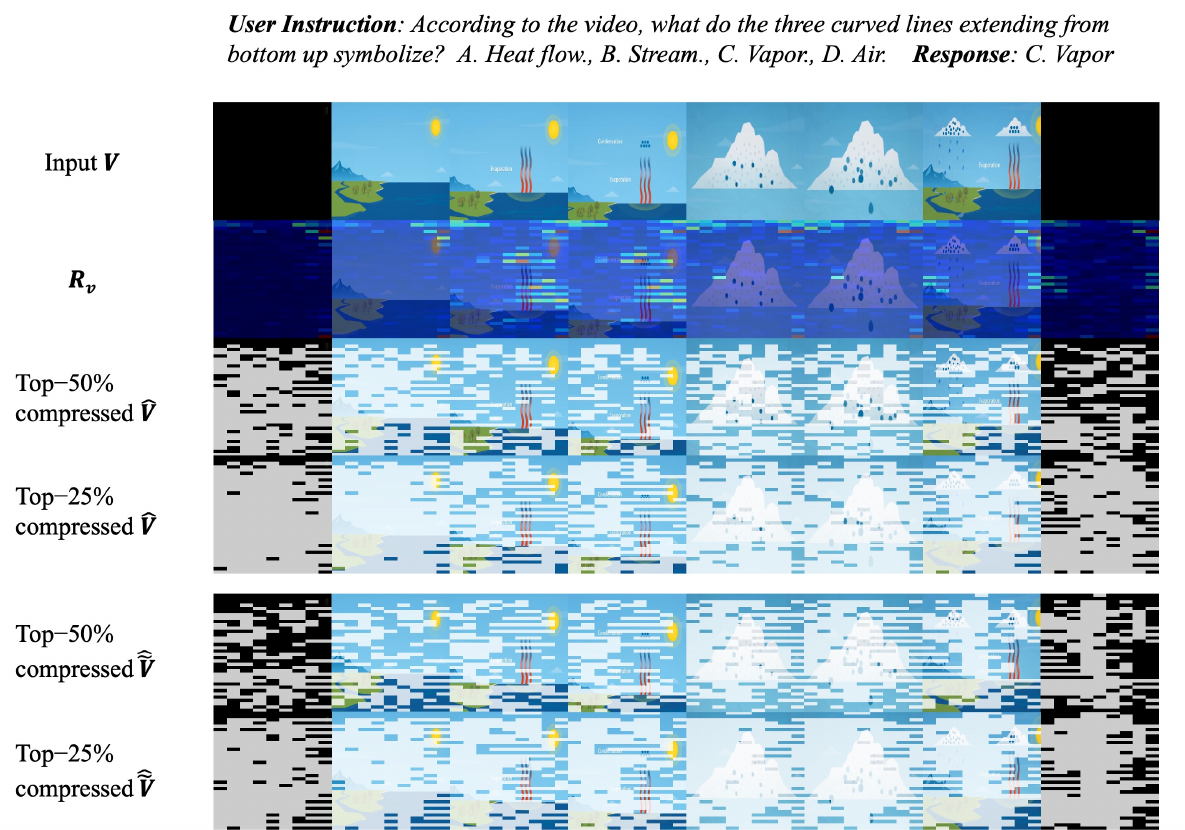}
  \caption{\textbf{Video Input Visualizations for Qwen2-VL}. }
  \label{fig:qwen_video}
\end{figure*}

\begin{figure*}[ht]
  \centering
  \includegraphics[width=\textwidth]{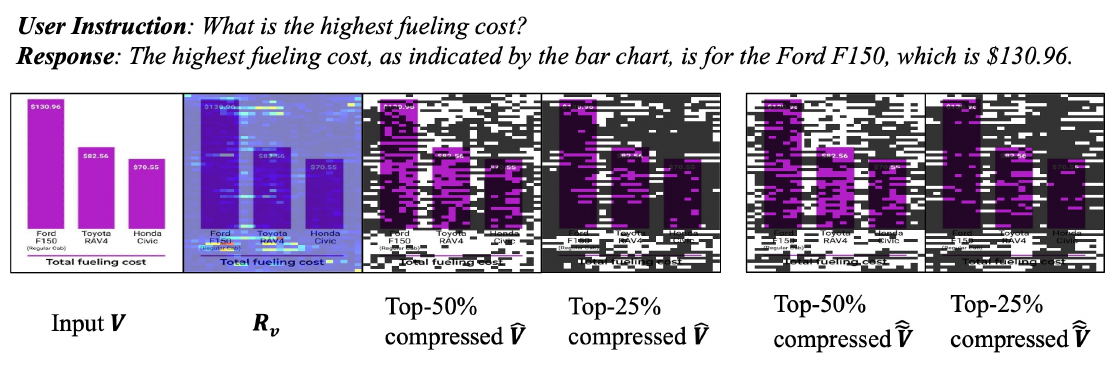}
  \caption{\textbf{Image Input Visualizations for Qwen2-VL}. }
  \label{fig:qwen_image}
\end{figure*}

\begin{figure*}[ht]
  \centering
  \includegraphics[width=\textwidth]{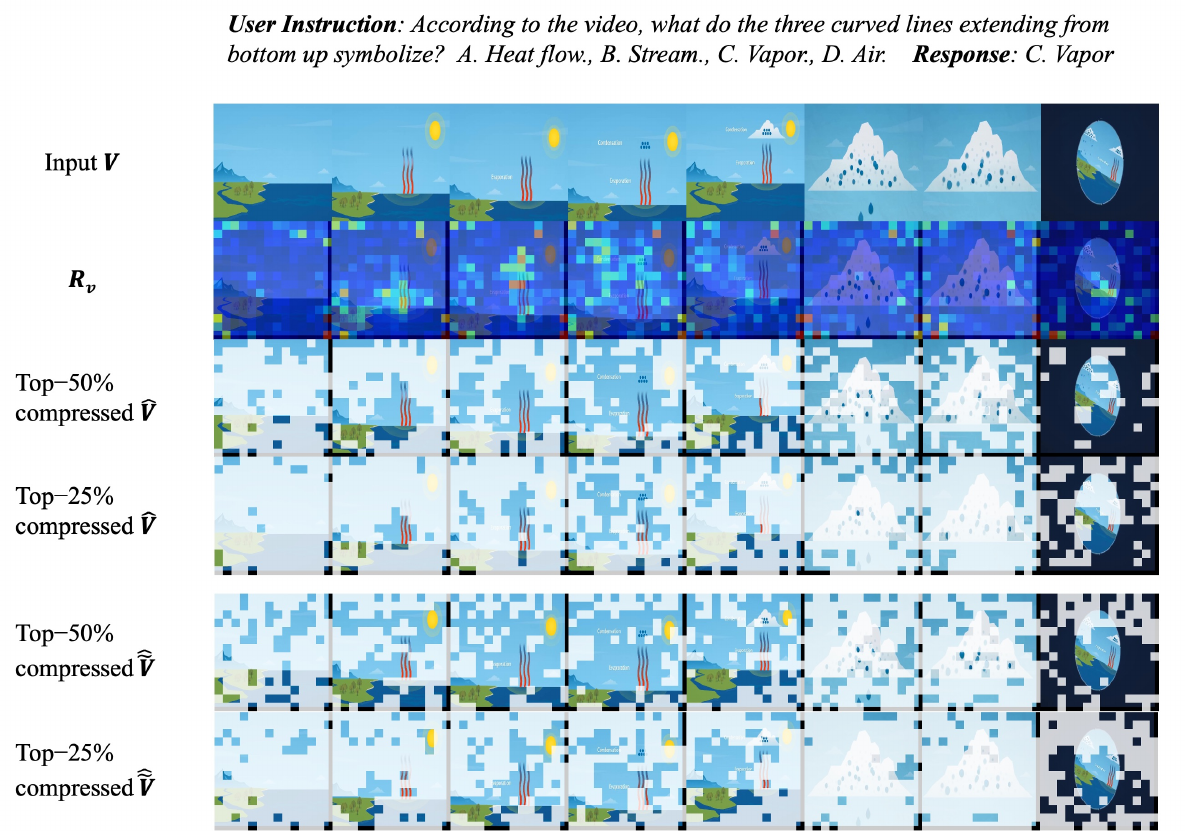}
  \caption{\textbf{Video Input Visualizations for VILA}. }
  \label{fig:vila_video}
\end{figure*}

We present visualization results for LLaVA-OneVision, Qwen2-VL, and VILA1.5 on both video and image inputs in Figures~\ref{fig:llavaov_video}-\ref{fig:vila_video}. Given an input image or video $V$, we first show the visual relevance scores $R_v$ with respect to the current response obtained using an explainability method. Based on $R_v$, we visualize the results of token pruning at 50\% and 25\% retention ratios (labeled as Top-50\% compressed $\hat{V}$ and Top-25\% compressed $\hat{V}$ in the figures). Then, we visualize the pruning results produced by our trained compressor ($f_\theta$) under the same compression ratios (labeled as Top-50\% compressed ${\scriptstyle\hat{\tilde{V}}}$ and Top-25\% compressed ${\scriptstyle\hat{\tilde{V}}}$ in the figures).

\subsection{Case Study: Explainability Reveals Instruction-Related Visual Tokens}
To demonstrate the effectiveness of explainability methods in identifying visual tokens that are highly relevant to instructions, we present 2 case studies covering both video and image inputs.

\begin{figure*}[ht]
  \centering
  \includegraphics[width=\textwidth]{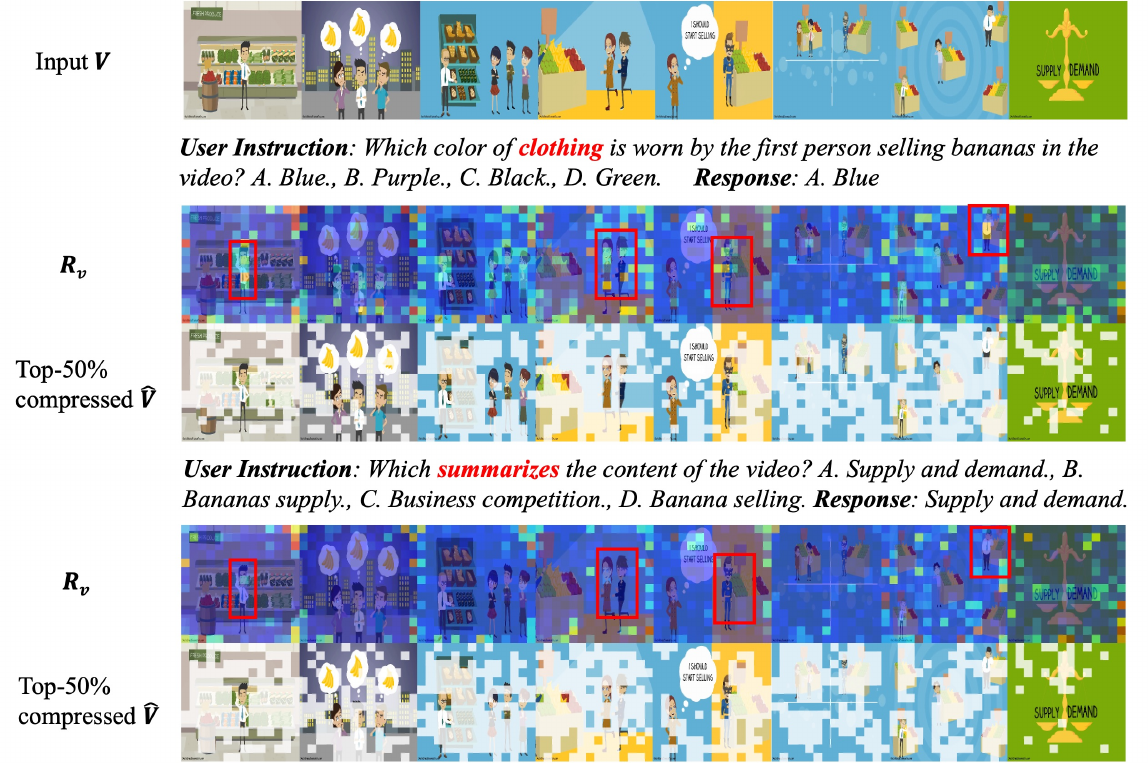}
  \caption{\textbf{Case Study 1}. }
  \label{fig:case_1}
\end{figure*}

\begin{figure*}[ht]
  \centering
  \includegraphics[width=\textwidth]{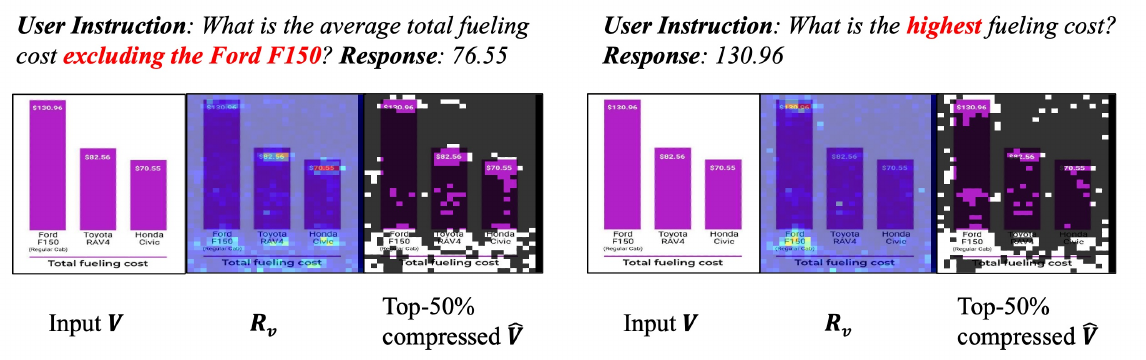}
  \caption{\textbf{Case Study 2}. }
  \label{fig:case_2}
\end{figure*}

Given the same input $V$, the explainability method generates visual relevance scores $R_v$ that selectively emphasize different visual tokens according to varying user instructions. As shown in Figure~\ref{fig:case_1}, when the user instruction specifically targets clothing-related information, the visual tokens corresponding to the person's clothing in the video obtain higher relevance scores compared to instructions requesting a general summary. Similarly, in Figure~\ref{fig:case_2}, visual tokens relevant to the user instruction exhibit higher relevance scores. When the user instruction specifies excluding the Ford F150, the visual attention shifts primarily to the other two columns. In contrast, when the instruction highlights the highest fueling cost, the Ford F150 column attracts nearly all the attention.

From a visualization standpoint, we further corroborate that the explanation results faithfully reflect the critical visual information required by the MLLM to answer the question.

\section{Implemantation Details.}
\label{app:2}
\noindent\textbf{Generating $R_v$}.
To derive $R_v$, our implementation employs eager attention, allowing access to full-layer attention maps required by the explainability method~\cite{gae_iccv}. Compared to FlashAttention~\cite{flashattn_nips} and inference based on KV cache~\cite{kv-cache}, eager attention requires more memory. To avoid out-of-memory errors and ensure efficient data generation, we limit the number of visual tokens to approximately 1500 per sample. Specifically, for video inputs, LLaVA-OneVision, VILA and Qwen2-VL are all set to sample 8 frames, resulting in 1569, 1568 and 1296 visual tokens per video, respectively. For image inputs, LLaVA-OneVision and Qwen2-VL use similar image resolutions, resulting in 1849 and 1500 visual tokens per image, respectively. VILA always processes an image as 196 tokens, eliminating the need for additional configuration. The generated $R_v$ can be used directly to guide token pruning or to train $f_\theta$.

\noindent\textbf{Training $f_\theta$}. $f_\theta$ is implemented as a five-layer fully convolutional network with channel dimensions of $32,64,128,256,$ and $512$. Each layer employs a 1D depthwise separable convolution~\cite{Xception}, \emph{i.e.}, a depthwise convolution with a kernel size of 3 followed by a pointwise convolution. An additional pointwise convolution layer is applied at the end for channel aggregation. The network is trained by using Adam~\cite{Adam} with default settings and a batch size of 128. Training data is collected from open-source datasets: a subset of LLaVA-Video~\cite{llava-video} for videos and a subset of Infinity-MM~\cite{Infinity-MM} for images, each containing approximately 10K samples. Note that $f_\theta$ is specific to MLLM, so each MLLM generates its own $A_v^0$ and $R_v$ based on the input image- or video-text pair for training.  The training is performed for roughly 100 epochs, taking about half an hour for image data and less than four hours for video data on a single A100 GPU.

\noindent\textbf{Details of Data for Training $f_\theta$.} We train our explainability-based compressor based on subsets sampled from high-quality open-source datasets. First, the details of the sampling are as follows:

\emph{Image Dataset.} For training the compressor used in image tasks, we sample a subset of Infinity-MM that ensures high quality and diversity. The training set primarily consists of data used during Stage 4, including 9k samples randomly sampled from the \textit{Data Generated by GPT-4} subset and 4k from \textit{Synthetic Data}.

\emph{Video Dataset.} For training the compressor used in video tasks, we sample a subset of LLaVA-Video. Specifically, we include 7k samples from \textit{LLaVA-Video}, 6k from \textit{NeXT-QA} and 4k from \textit{ActivityNetQA}. Note that the training sets of \textit{NeXT-QA} and \textit{ActivityNetQA} have no overlap with the testing sets used in the evaluation. During sampling, since LLaVA-Video contains several parts categorized by task type (open-ended and multi-choice) and video duration (0–30s, 30-60s, 1–2min and 2-3min), we ensure a balanced distribution by randomly selecting an equal number of training examples from each part.

Moreover, we assume that the visual attention distributions ($R_v$) associated with correct answers exhibit higher quality than those that lead to incorrect answers. Therefore, when training $f_\theta$ for a specific MLLM, the sampled data are evaluated by this MLLM, and the samples with incorrect answers are filtered out. Only samples for which the MLLM produces correct answers are retained and used as training data. The number of the retained samples ranges from 8K to 12K.

\noindent\textbf{Inference}. 
The learned $f_\theta$ can be seamlessly integrated into existing inference pipelines (no modifications are required for the prefill and decode phases of LLM inference). More interestingly, $f_\theta$ is capable of processing longer $A_v^0$ thanks to the fully convolution design. That is, our compression method can handle larger images and longer videos, even though the visual token number is limited to approximately 1500 during training. Corresponding experiments have been conducted. In these experiments, Qwen2-VL dynamically processes both images (with `max\textunderscore pixels' set to half of its default value) and videos (with `VIDEO\textunderscore MAX\textunderscore PIXELS' and `FPS\textunderscore MAX\textunderscore FRAMES' set to 384$\times$28$\times$28 and 32, respectively). These configurations are set to accommodate hardware~resource constraints. LLaVA-OneVision also processes images dynamically with default settings, while sampling 32 frames per video as in~\cite{prunevid_arxiv} for a fair comparison. For VILA, the input image size cannot be changed, and the number of input video frames is set to 16. 

\section{Efficiency Analysis in Inference}
\label{app:3}
To evaluate computational efficiency during inference, we report the FLOPs of the visual token part. Specifically, we consider the FLOPs of the multihead attention and the feed-forward network (FFN) modules as:
% We follow  in using the FLOPs of the visual token part to evaluate inference-time efficiency. Specifically, we consider the FLOPs of the multihead attention and the feed-forward network modules as:
\begin{equation}
    \mathrm{FLOPs_{layer}} = 4nd^2+2n^2d+lnm ,
\label{eq:one_layer}
\end{equation}
where $n$ is the number of visual tokens, $d$ is the hidden state size, $m$ is the intermediate size of the FFN, and $l$ is the number of layers in the FFN. To compute the total FLOPs for the entire LLM, we simply multiply Eq.~\ref{eq:one_layer} by the number of Transformer layers $N_L$, i.e., $\mathrm{FLOPs_{LLM}} = N_L(4nd^2+2n^2d+lnm)$.

At the input stage of the LLM, our compressor introduces additional computation. First, we consider the FLOPs introduced by the first-layer attention map:
\begin{equation}
    \mathrm{FLOPs_{attn}} = nd^2+nd .
\label{eq:attention_map}
\end{equation}
Note that only the key projection computation for visual tokens and the attention computation from textual tokens to visual tokens are required, corresponding to the term $nd^2$ and $nd$, respectively. Only the FLOPs incurred by the visual part are included.
% Since we only consider the FLOPs incurred by the visual part, the computation related to the user instruction is omitted from the calculation. The term $nd^2$ corresponds to the computation of the key projection for the visual tokens and $nd$ represents the FLOPs for computing the attention map through the query-key dot product.

Next, we account for the FLOPs introduced by the 1D depthwise separable convolution:
\begin{equation}
    \mathrm{FLOPs_{conv}} = \sum_{l=1}^{L} n(C_{in}^{l}k+C_{in}^{l}C_{out}^{i}) ,
\label{eq:conv}
\end{equation}
where $C_{in}^{l}$ and $C_{out}^{l}$ denote the number of input and output channels of the l-th layer, respectively. We ensure that the output shape of each convolutional layer remains the same as its input by applying appropriate padding with respect to the kernel size $k$. As a result, the number of visual tokens $n$ 
remains constant across all layers. Then the total FLOPs is computed as the sum of the operations across all $L$ convolutional layers.

To intuitively understand the additional computational cost introduced by our method, we adopt a typical parameter configuration used in MLLMs. Specifically, we set the number of visual tokens $n$ to 1568, the hidden dimension $d$ to 3584, the intermediate size $m$ to 18944, and assume 3 layers per FFN block ($l = 3$). For the full LLM, we consider a 28-layer Transformer blocks ($N_L = 28$). For $f_\theta$, we follow the configuration described in Section 4.1 (Experimental Setup). Concretely, the convolutional network consists of 5 layers ($L = 5$) with kernel size $k = 3$, and channel dimensions increasing across layers: 32, 64, 128, 256, and 512. Based on these settings, $\mathrm{FLOPs_{attn}}$ amounts to approximately 0.02 trillion, $\mathrm{FLOPs_{conv}}$ is approximately 0.0003 trillion, while $\mathrm{FLOPs_{LLM}}$ reaches approximately 11.69 trillion. It can be observed that the computational overhead introduced by our compressor is negligible. The computational costs of these two parts account for only \textbf{0.17\%} and \textbf{0.0026\%} of the total computational cost, respectively.

The FLOPs reported in the Table~\ref{tab:image_base}, Table~\ref{tab:video_base}, Table~\ref{tab:efficiency} and Figure~\ref{fig:all} are computed using a standardized input setting. For image input, FLOPs are computed using a 384 $\times$ 512 input image as the reference (the number of visual tokens $n$ is 1728 for LLaVA-OneVision and 1302 for Qwen2-VL). For video input, LLaVA-OneVision and VILA1.5 sample 32 and 16 frames, respectively, resulting in visual token counts $n$ of 6272 and 3136. We fix Qwen2-VL's input to 32 frames at 720 $\times$ 1280 resolution ($n$=5824) for FLOPs calculation.

\section{Additional Ablation Results}
\label{app:6}
We perform additional ablation studies to validate the design choices of our method, specifically examining (i) the depth of the convolutional network and (ii) the attention layer index used as input. For the convolutional depth, we conduct controlled experiments on Qwen2-VL using networks of 3, 5, 7, and 10 layers, while keeping all other factors fixed, including the training data, input representation $A^0$, and training configurations (learning rate, batch size, and number of epochs). All variants follow a consistent channel-growth strategy in which the network starts with 32 channels and gradually increases capacity, with intermediate channel dimensions repeated to moderate growth and avoid over-parameterization. The exact configurations are 3-layer: [32, 64, 128], 5-layer: [32, 64, 128, 256, 512], 7-layer: [32, 64, 128, 128, 256, 256, 512], and 10-layer: [32, 64, 128, 128, 256, 256, 512, 512, 512, 512]. For the study on attention-layer index, we train our lightweight model on Qwen2-VL using attention maps extracted from different transformer layers as input, specifically evaluating layers 1, 2, 4, and 6.

The results are reported in Table~\ref{tab:ablation_cnn_layers} and Table~\ref{tab:ablation_input_layer}, respectively. Table~\ref{tab:ablation_cnn_layers} shows that the 5-layer model achieves the best performance under retention ratios of 50\% and 25\%. Increasing the depth to 7 or 10 layers leads to a slight performance degradation—especially under the more challenging 25\% retention condition. This suggests that deeper networks begin to overfit or suffer from optimization difficulties given the simplicity of the task. In contrast, the 3-layer model lacks sufficient capacity to capture the relevance distribution. Therefore, the 5-layer architecture strikes the optimal balance between model capacity and task complexity, which justifies our design choice.

Table~\ref{tab:ablation_input_layer} shows that using multi-layer attention scores as input yields only marginal gains—+0.1\% average at 50\% retention (with 2 layers) and +0.6\% at 25\% retention (with 4 layers)—yet incurs substantial computational overhead, with FLOPs increasing from 0.24$\times$ to 0.43$\times$ at 25\% retention, almost two times. Notably, the first-layer input achieves nearly the same performance as multi-layer variants while being significantly more efficient.
Our work aims to achieve an innovative paradigm shift that enables task-related token compression to be applied prior to the LLM, which not only significantly reduces computation and memory overhead during both prefill and decode phases, but also allows deployment without modifying the LLM architecture. However, leveraging deeper-layer attention maps as input yields only marginal improvements while contradicting our design goal. Overall, the first-layer attention map suffices for our purpose.

\begin{table*}[t]
    \caption{\textbf{Ablation study on the depth of the convolutional network .} In our work, we use a 5-layer convolutional network as the explainability-based compressor before the LLM.}
    \vspace{8pt}
    \renewcommand{\arraystretch}{1.4}
    \setlength{\tabcolsep}{3pt}
    \centering
    \resizebox{1\textwidth}{!}{
    \begin{tabular}{
                >{\arraybackslash}p{2cm}
                |c|
                >{\centering\arraybackslash}p{1.5cm}|
                >{\centering\arraybackslash}p{1.5cm}|
                >{\centering\arraybackslash}p{1.5cm}
                >{\centering\arraybackslash}p{1.5cm}
                >{\centering\arraybackslash}p{1.5cm}|
                >{\centering\arraybackslash}p{2cm}
                >{\centering\arraybackslash}p{2cm}
                >{\centering\arraybackslash}p{2cm}|c}
        \toprule
        \multirow{2}{*}{\textbf{Methods}} & \multirow{2}{*}{\parbox[c]{1.5cm}{\centering \textbf{Conv Depth}}} & \multirow{2}{*}{\parbox[c]{1.5cm}{\centering \textbf{Retention Ratio}}} & \multirow{2}{*}{\textbf{FLOPs}} & \multicolumn{3}{c|}{\textbf{Image Benchmark}}& \multicolumn{3}{c|}{\textbf{Video Benchmark}} & \multirow{2}{*}{\textbf{Avg.(\%)}}\\
        \cline{5-7} \cline{8-10}
        & & & & \textbf{MME} & \textbf{MMStar} & \textbf{MMVet} & \textbf{Video-MME} & \textbf{MVBench} & \textbf{MMBench-V} & \\
        \hline
        \cellcolor{gray!20}Qwen2-VL & \cellcolor{gray!20}- & \cellcolor{gray!20}100\% & \cellcolor{gray!20}1.00$\times$ & \cellcolor{gray!20} 2295.1 & \cellcolor{gray!20}60.4 & \cellcolor{gray!20}54.0 & \cellcolor{gray!20}50.4 & \cellcolor{gray!20}51.0 & \cellcolor{gray!20}1.23 & \cellcolor{gray!20}100\\

        \hline
        \multirow{8}{*}{\parbox[c]{2cm}{\textbf{Qwen2-VL w/ Ours}}}& 3 & 50\% & 0.49$\times$ & 2265.0 & 55.5 & 50.4 & 49.8 & 49.3 & 1.15 & 95.5 \\
        
        & \textbf{5} & 50\% & 0.49$\times$ &2288.3 & 55.9 & 51.9 & 50.0 & 49.8 & 1.18 & \textbf{96.9} \\
        & 7 & 50\% & 0.49$\times$ &2279.4 & 56.1 & 51.9 & 49.8 & 49.4 & 1.17 & 96.5 \\
        & 10 & 50\% & 0.49$\times$ &2275.1 & 55.1 & 51.5 & 50.0 & 49.7 & 1.15 & 96.0 \\

        \cline{2-11}
        & 3 & 25\% & 0.24$\times$ & 2297.1 & 60.3 & 53.2 & 51.0 & 50.7 & 1.19 & 90.7 \\
        & \textbf{5} & 25\% & 0.24$\times$ & 2299.1 & 58.7 & 51.7 & 50.3 & 49.7 & 1.17 & \textbf{91.7} \\
        & 7 & 25\% & 0.24$\times$ & 2299.1 & 58.7 & 51.7 & 50.3 & 49.7 & 1.17 & 91.1 \\
        & 10 & 25\% & 0.24$\times$ & 2299.1 & 58.7 & 51.7 & 50.3 & 49.7 & 1.17 & 90.4 \\
    
        \bottomrule
    \end{tabular}
    }
    \label{tab:ablation_cnn_layers}
\end{table*}

\begin{table*}[t]
    \caption{\textbf{Ablation study on the attention layer index used as input.} In our work, the explainability-based compressor takes the first-layer attention map $A^0$ as its input.}
    \vspace{8pt}
    \renewcommand{\arraystretch}{1.4}
    \setlength{\tabcolsep}{3pt}
    \centering
    \resizebox{1\textwidth}{!}{
    \begin{tabular}{
                >{\arraybackslash}p{2cm}
                |c|
                >{\centering\arraybackslash}p{1.5cm}|
                >{\centering\arraybackslash}p{1.5cm}|
                >{\centering\arraybackslash}p{1.5cm}
                >{\centering\arraybackslash}p{1.5cm}
                >{\centering\arraybackslash}p{1.5cm}|
                >{\centering\arraybackslash}p{2cm}
                >{\centering\arraybackslash}p{2cm}
                >{\centering\arraybackslash}p{2cm}|c}
        \toprule
        \multirow{2}{*}{\textbf{Methods}} & \multirow{2}{*}{\parbox[c]{1.5cm}{\centering \textbf{Input Layer}}} & \multirow{2}{*}{\parbox[c]{1.5cm}{\centering \textbf{Retention Ratio}}} & \multirow{2}{*}{\textbf{FLOPs}} & \multicolumn{3}{c|}{\textbf{Image Benchmark}} & \multicolumn{3}{c|}{\textbf{Video Benchmark}} & \multirow{2}{*}{\textbf{Avg.(\%)}}\\
        \cline{5-7} \cline{8-10}
        & & & & \textbf{MME} & \textbf{MMStar} & \textbf{MMVet} & \textbf{Video-MME} & \textbf{MVBench} & \textbf{MMBench-V} & \\
        \hline
        \cellcolor{gray!20}Qwen2-VL & \cellcolor{gray!20}- & \cellcolor{gray!20}100\% & \cellcolor{gray!20}1.00$\times$ & \cellcolor{gray!20} 2295.1 & \cellcolor{gray!20}60.4 & \cellcolor{gray!20}54.0 & \cellcolor{gray!20}50.4 & \cellcolor{gray!20}51.0 & \cellcolor{gray!20}1.23 & \cellcolor{gray!20}100\\

        \hline
        \multirow{8}{*}{\parbox[c]{2cm}{\textbf{Qwen2-VL w/ Ours}}}& \textbf{1} & 50\% & \textbf{0.49$\times$} & 2288.3 & 55.9 & 51.9 & 50.0 & 49.8 & 1.18 & 96.9 \\
        
        & 2 & 50\% & 0.53$\times$ &2279.8 & 55.8 & 52.0 & 50.1 & 50.2 & 1.18 & \textbf{97.0} \\
        & 4 & 50\% & 0.60$\times$ &2303.0 & 56.3 & 52.2 & 49.3 & 50.1 & 1.15 & 96.6 \\
        & 6 & 50\% & 0.67$\times$ &2278.1 & 56.7 & 49.5 & 49.4 & 49.6 & 1.19 & 96.1 \\

        \cline{2-11}
        & \textbf{1} & 25\% & \textbf{0.24$\times$} & 2280.9 & 51.8 & 47.3 & 48.1 & 46.7 & 1.11 & 91.7 \\
        & 2 & 25\% & 0.35$\times$ & 2272.0 & 51.9 & 46.6 & 47.8 & 48.4 & 1.10 & 91.7 \\
        & 4 & 25\% & 0.35$\times$ & 2270.2 & 52.1 & 48.0 & 48.0 & 48.1 & 1.11 & \textbf{92.3} \\
        & 6 & 25\% & 0.43$\times$ & 2252.3 & 52.1 & 47.2 & 47.4 & 47.6 & 1.11 & 91.6 \\
    
        \bottomrule
    \end{tabular}
    }
    \label{tab:ablation_input_layer}
\end{table*}

\section{Additional Experimental Results}
\label{app:4}

For image tasks, we further conducted evaluations across additional benchmarks, including GQA~\cite{GQA} (real-world visual reasoning), ScienceQA (SQA)~\cite{SQA} (scientific reasoning) , VizWiz~\cite{VizWiz} (real-world robustness), ChartQA~\cite{chartqa} (chart reasoning), DocVQA~\cite{docvqa} (document reasoning), OCRBench~\cite{ocrbench} (OCR reasoning), to comprehensively assess the effectiveness of our method. The results are presented in Table~\ref{tab:image_base_add}, our method consistently outperforms the baselines across diverse image benchmarks, including real-world visual reasoning and OCR-related tasks.

For video tasks, we also investigate a more aggressive compression setting with a 10\% retention ratio in Table~\ref{tab:video_base_add}, as a supplement to Table~\ref{tab:video_base}. Our method attains the lowest FLOPs while preserving competitive accuracy, achieving average improvements of 3.7\%, 3.2\%, and 1.9\% across all benchmarks for LLaVA-OneVision, Qwen2-VL, and VILA, respectively, compared to the best-performing baseline. Notably, even under such extreme compression, our method consistently delivers strong results, highlighting its robustness across different MLLMs.

We provide full tables of results corresponding to the generalization experiments shown in the Figure~\ref{fig:all} (a) in the main text (Applying to Larger Images and Longer Videos), with detailed results for the image and video benchmarks listed in Table~\ref{tab:image_extend} and Table~\ref{tab:video_extend}, respectively. In addition, we provide detailed comparison results shown in the Figure~\ref{fig:all} (b) for two challenging benchmarks, MMStar and MVBench, in Table~\ref{tab:mmstar} and Table~\ref{tab:mvbench}.

Table~\ref{tab:efficiency2} exhibits the additional efficiency analysis on MMVet. Our lightweight compressor achieves substantial reductions in KV-cache usage and accelerates the prefill stage, while achieving the highest task scores and keeping overall inference time comparable to baselines. These results demonstrate that our approach maintains both strong task performance and computational efficiency.

\begin{table*}
    \caption{\textbf{Compare explainability-based compressor on image benchmarks.} Values marked with * in Retention Ratio denote the average retention ratio across LLM layers due to multi-stage compression in PDrop.}
    \vspace{8pt}
    \renewcommand{\arraystretch}{1.3}
    \setlength{\tabcolsep}{15pt}
    \centering
    \resizebox{1\textwidth}{!}{
    \begin{tabular}{l|c|c|cccccc}
        \toprule
        \multirow{2}{*}{\makebox[0.005\textwidth][l]{\textbf{Method}}} &
        \makebox[0.05\textwidth][c]{\textbf{Retention}} &
        \multirow{2}{*}{\makebox[0.005\textwidth][c]{\textbf{FLOPs}}} &
        \multirow{2}{*}{\makebox[0.005\textwidth][c]{\textbf{GQA}}} &
        \multirow{2}{*}{\makebox[0.005\textwidth][c]{\textbf{SQA}}} &
        \multirow{2}{*}{\makebox[0.005\textwidth][c]{\textbf{VizWiz}}} &
        \multirow{2}{*}{\makebox[0.005\textwidth][c]{\textbf{ChartQA}}} &
        \multirow{2}{*}{\makebox[0.005\textwidth][c]{\textbf{DocVQA}}} &
        \multirow{2}{*}{\makebox[0.005\textwidth][c]{\textbf{OCRBench}}}\\

        &\makebox[0.05\textwidth][c]{\textbf{Ratio}}& & & & & & &\\
        \hline
        \cellcolor{gray!20}LLaVA-OneVision & \cellcolor{gray!20}100\% & \cellcolor{gray!20}1.00$\times$& \cellcolor{gray!20}63.1 & \cellcolor{gray!20}95.1 & \cellcolor{gray!20}33.1 & \cellcolor{gray!20}71.4 & \cellcolor{gray!20}73.3& \cellcolor{gray!20}54.4\\
        
        LLaVA-OneVision w/ FastV & 50\% & 0.51$\times$ & 42.5 & 76.2 & 3.2 & 41.4 & 42.9 & 28.3 \\
        
        LLaVA-OneVision w/ Pdrop & 51\%* & 0.51$\times$ & 61.0 & \underline{92.0} & 29.7 & \underline{62.0} & \underline{58.0} & 42.0\\
        
        LLaVA-OneVision w/ Dart & 50\% & 0.51$\times$ & \textbf{61.7} & 91.7 & \underline{29.8} & 61.0 & 55.3 & \underline{43.2} \\
        
        \textbf{LLaVA-OneVision w/ Ours} & 50\% & \textbf{0.48}\boldsymbol{$\times$} & \underline{61.4} & \textbf{92.2}  & \textbf{33.6} & \textbf{63.0} & \textbf{62.0} & \textbf{47.0}\\

        \hline
        LLaVA-OneVision w/ FastV & 25\% & 0.27$\times$ & 39.9 & 70.4 & 2.7 & 21.3 & 22.2 & 13.9 \\
        
        LLaVA-OneVision w/ PDrop & 25\%* & 0.25$\times$ & 57.2 & 87.9 & 25.3 & 43.4 & 34.9 & 30.2\\
        
        LLaVA-OneVision w/ Dart & 25\% & 0.27$\times$ & \underline{57.5} & \underline{89.1} & \underline{27.9 }& \underline{46.7} & \underline{39.9} & \underline{34.3}\\
        
        \textbf{LLaVA-OneVision w/ Ours} & 25\% & \textbf{0.24}\boldsymbol{$\times$} & \textbf{59.0} & \textbf{90.2} & \textbf{33.0} & \textbf{52.7} & \textbf{49.4} & \textbf{38.3}\\
        
        \hline
        \cellcolor{gray!20}Qwen2-VL & \cellcolor{gray!20}100\% & \cellcolor{gray!20}1.00$\times$ & \cellcolor{gray!20}62.2 & \cellcolor{gray!20}85.7 & \cellcolor{gray!20}44.3 & \cellcolor{gray!20}92.7 & \cellcolor{gray!20}93.1 & \cellcolor{gray!20}81.5\\
        
        Qwen2-VL w/ FastV & 50\% & 0.51$\times$ & 43.9 & 65.3 & 32.5 & 51.3 & 59.2 & 56.3\\
        
        Qwen2-VL w/ PDrop & 51\%* & 0.51$\times$ & 60.8 & 83.5 & 41.9 & 77.0 & 78.5 & \underline{78.8}\\ 
        
        Qwen2-VL w/ Dart & 50\% & 0.51$\times$ & \underline{61.3} & \underline{84.7} & \underline{43.9} & \underline{78.4} & \underline{79.7} & 78.5\\
        
        \textbf{Qwen2-VL w/ Ours} & 50\% & \textbf{0.49}\boldsymbol{$\times$} & \textbf{61.6} & \textbf{85.5} & \textbf{44.1} & \textbf{80.1} & \textbf{81.2} & \textbf{79.5}\\
        
        \hline
        Qwen2-VL w/ FastV & 25\% & 0.27$\times$ & 41.5 & 64.3 & 31.5 & 41.1 & 46.1 & 46.8\\
        
        Qwen2-VL w/ PDrop & 25\%* & 0.25$\times$ & 57.1 & 83.2 & 40.8 & 61.7 & 58.2 & 63.9\\
        
        Qwen2-VL w/ Dart & 25\% & 0.27$\times$ & \underline{58.8} & \underline{83.9} & \underline{41.3} & \underline{65.0} & \underline{62.8} & \underline{65.1}\\
        
        \textbf{Qwen2-VL w/ Ours} & 25\% & \textbf{0.24}\boldsymbol{$\times$} & \textbf{59.3} & \textbf{84.3} & \textbf{43.1} & \textbf{70.9} & \textbf{72.1} & \textbf{67.8}\\
        
        \bottomrule
    \end{tabular}
    }
    \label{tab:image_base_add}
\end{table*}

\begin{table*}
    \caption{\textbf{Compare explainability-based compressor on video benchmarks.}}
    \vspace{8pt}
    \renewcommand{\arraystretch}{1.3}
    \setlength{\tabcolsep}{1pt}
    \centering
    \resizebox{1\textwidth}{!}{
    \begin{tabular}{l|c|c|cccccc|c}
        \toprule
        % 可以加一行type
        \multirow{2}{*}{\makebox[0.2\textwidth][l]{\textbf{Method}}}&
        \multirow{2}{*}{\parbox[c]{1.5cm}{\centering \textbf{Retention Ratio}}}& 
        \multirow{2}{*}{\makebox[0.1\textwidth][c]{\textbf{FLOPs}}} &
        \multirow{2}{*}{\makebox[0.15\textwidth][c]{\textbf{Video-MME}}} &
        \multirow{2}{*}{\makebox[0.15\textwidth][c]{\textbf{MVBench}}} &
        \multirow{2}{*}{\parbox[c]{1.5cm}{\centering \textbf{MMBench-Video}}} &
        \multicolumn{2}{c}{\makebox[0.2\textwidth][c]{\textbf{Next-QA}}} &
        \multirow{2}{*}{\makebox[0.15\textwidth][c]{\textbf{Activity-QA}}} &
        \multirow{2}{*}{\makebox[0.1\textwidth][c]{\textbf{Avg.(\%)}}}\\
        \cline{7-8} 
        & & & & & & \textbf{multi-choice} & \textbf{open-ended} & & \\
        \hline
        \cellcolor{gray!20}LLaVA-OV & \cellcolor{gray!20}100\% & \cellcolor{gray!20}1.00$\times$ & \cellcolor{gray!20}53.6 & \cellcolor{gray!20}41.2 & \cellcolor{gray!20}0.41 & \cellcolor{gray!20}79.2 & \cellcolor{gray!20}49.0 & \cellcolor{gray!20}56.9 & \cellcolor{gray!20}100\\
        
        LLaVA-OV w/ FastV & 10\% & 0.12$\times$ & 37.7 & 22.4 & 0.27 & 60.3 & 30.6 &39.9 & 66.5\\
        LLaVA-OV w/ PDrop & 10\%* & 0.10$\times$ & 47.0 & 37.0 & 0.35 & 72.3 & \underline{43.7} & 50.0 & 88.5 \\
        LLaVA-OV w/ Dart & 10\% & 0.12$\times$ & \textbf{47.3} & \underline{36.9} & \underline{0.36} & \underline{72.7} & 43.5 & \underline{50.3} & 89.1 \\
        \textbf{LLaVA-OV w/ Ours} & 10\% & 0.09$\times$ & \underline{47.1} & \textbf{37.4} & \textbf{0.40} & \textbf{76.5} & \textbf{45.6} & \textbf{51.6} & \textbf{92.8} \\
        
        \hline
        \cellcolor{gray!20}Qwen2-VL & \cellcolor{gray!20}100\% & \cellcolor{gray!20}1.00$\times$ & \cellcolor{gray!20}50.4 & \cellcolor{gray!20}51.0 & \cellcolor{gray!20}1.23 & \cellcolor{gray!20}76.8 & \cellcolor{gray!20}45.5 & \cellcolor{gray!20}53.6 & \cellcolor{gray!20}100\\
        
        Qwen2-VL w/ FastV & 10\% & 0.12$\times$ & 29.1 & 37.5 & 0.44 & 39.4 & 23.3 & 32.0 & 54.9 \\
        Qwen2-VL w/ PDrop & 10\%* & 0.10$\times$ & 45.2 & 40.3 & 0.82 & 71.5 & 41.8 & 45.1 & 84.1 \\
        Qwen2-VL w/ Dart & 10\% & 0.12$\times$ & \underline{45.8} & \underline{41.6} & \underline{0.85} & \textbf{72.1} & \underline{42.6} & \underline{45.5} & 85.7 \\
        \textbf{Qwen2-VL w/ Ours} & 10\% & 0.09$\times$ & \textbf{46.1} & \textbf{42.5} & \textbf{1.00} & \underline{72.0} & \textbf{43.3} & \textbf{47.5} & \textbf{88.9} \\
        
        \hline
        \cellcolor{gray!20}VILA & \cellcolor{gray!20}100\% & \cellcolor{gray!20}1.00$\times$ & \cellcolor{gray!20}47.3 & \cellcolor{gray!20}34.0 & \cellcolor{gray!20}1.29 & \cellcolor{gray!20}69.9 & \cellcolor{gray!20}46.2 & \cellcolor{gray!20}55.6 & \cellcolor{gray!20}100\\
        
        VILA w/ FastV & 10\% & 0.12$\times$ & 37.8 & 19.5 & 0.88 & 57.9 & 33.9 &43.2 & 73.2 \\
        VILA w/ PDrop & 10\%* & 0.11$\times$ & \underline{43.0} & 34.4 & 1.13 & \underline{65.2} & 43.5 & 50.6 & 93.0 \\
        VILA w/ Dart & 10\% & 0.12$\times$ & 42.9 & \underline{34.8} & \underline{1.14} & 64.8 & \underline{43.7} & \underline{51.1} & 93.4 \\
        \textbf{VILA w/ Ours} & 10\% & 0.09$\times$ & \textbf{43.6} & \textbf{35.0} & \textbf{1.14} & \textbf{67.0} & \textbf{44.7} & \textbf{53.0} & \textbf{95.3} \\
        
        \bottomrule
    \end{tabular}}
    \label{tab:video_base_add}
\end{table*}

\begin{table*}
    \caption{\textbf{Compare generalization performance of our compressor on image benmarks.}}
    \vspace{8pt}
    \renewcommand{\arraystretch}{1.3}
    \setlength{\tabcolsep}{15pt}
    \centering
    \resizebox{1\textwidth}{!}{
    \begin{tabular}{l|c|c|cccc|c}
        \toprule
        \multirow{2}{*}{\makebox[0.12\textwidth][l]{\textbf{Method}}} &
        \makebox[0.05\textwidth][c]{\textbf{Rentention}} &
        \multirow{2}{*}{\makebox[0.05\textwidth][c]{\textbf{FLOPs}}} &
        \multirow{2}{*}{\makebox[0.005\textwidth][c]{\textbf{MME}}} &
        \multirow{2}{*}{\makebox[0.005\textwidth][c]{\textbf{MMStar}}} &
        \multirow{2}{*}{\makebox[0.005\textwidth][c]{\textbf{MMVet}}} &
        \multirow{2}{*}{\makebox[0.005\textwidth][c]{\textbf{SEED}}} &
        \multirow{2}{*}{\makebox[0.0005\textwidth][c]{\textbf{Avg.(\%)}}} \\

        &\makebox[0.05\textwidth][c]{\textbf{Ratio}}& & & & & &\\
        \hline
        \cellcolor{gray!20}LLaVA-OneVision & \cellcolor{gray!20}100\% & \cellcolor{gray!20}1.00$\times$& \cellcolor{gray!20}2002.0 & \cellcolor{gray!20}62.0 & \cellcolor{gray!20}52.0 & \cellcolor{gray!20}76.7 & \cellcolor{gray!20}100 \\
        
        LLaVA-OneVision w/ FastV\textsubscript{Opt.Config.} & 50\% & 0.51$\times$ & 1990.3 & 57.3 & 48.4 & 75.7 & 95.9\\
        \textbf{LLaVA-OneVision w/ Ours} & 50\% & 0.48$\times$ & 1988.0 & 57.8 & 50.2 & 75.4 & \textbf{96.8}\\

        \hline
        LLaVA-OneVision w/ FastV\textsubscript{Opt.Config.} & 25\% & 0.27$\times$ & 1953.7 & 52.0 & 43.2 & 71.5 & 89.4\\
        \textbf{LLaVA-OneVision w/ Ours} & 25\% & 0.24$\times$ & 1985.8 & 52.6 & 45.8 & 73.1 & \textbf{91.9}\\
        
        \hline
        \cellcolor{gray!20}Qwen2-VL & \cellcolor{gray!20}100\% & \cellcolor{gray!20}1.00$\times$ & \cellcolor{gray!20}2316.6 & \cellcolor{gray!20}61.1 & \cellcolor{gray!20}51.7 & \cellcolor{gray!20}76.4 & \cellcolor{gray!20}100 \\
        
        Qwen2-VL w/ FastV\textsubscript{Opt.Config.} & 50\% & 0.51$\times$ & 2295.8 & 57.7 & 52.4 & 74.8 & 98.2\\
        \textbf{Qwen2-VL w/ Ours} & 50\% & 0.49$\times$ & 2311.7 & 57.9 & 53.9 & 73.9 & \textbf{98.9} \\
        \hline
        Qwen2-VL w/ FastV\textsubscript{Opt.Config.} & 25\% & 0.27$\times$ & 2288.2 & 55.0 & 49.3 & 71.1 &94.3\\
        \textbf{Qwen2-VL w/ Ours} & 25\% & 0.24$\times$ & 2283.1 & 55.8 & 50.8 & 71.0 & \textbf{95.3}\\
        \bottomrule
    \end{tabular}
    }
    \label{tab:image_extend}
\end{table*}

\begin{table*}
    \caption{\textbf{Compare generalization performance of our compressor on video benchmarks.}}
    \vspace{8pt}
    \renewcommand{\arraystretch}{1.3}
    \setlength{\tabcolsep}{1pt}
    \centering
    \resizebox{1\textwidth}{!}{
    \begin{tabular}{l|c|c|cccccc|c}
        \toprule
        % 可以加一行type
        \multirow{2}{*}{\makebox[0.2\textwidth][l]{\textbf{Method}}}&
        \multirow{2}{*}{\parbox[c]{1.5cm}{\centering \textbf{Retention Ratio}}}& 
        \multirow{2}{*}{\makebox[0.1\textwidth][c]{\textbf{FLOPs}}} &
        \multirow{2}{*}{\makebox[0.15\textwidth][c]{\textbf{Video-MME}}} &
        \multirow{2}{*}{\makebox[0.15\textwidth][c]{\textbf{MVBench}}} &
        \multirow{2}{*}{\parbox[c]{1.8cm}{\centering \textbf{MMBench-Video}}} &
        \multicolumn{2}{c}{\makebox[0.12\textwidth][c]{\textbf{Next-QA}}} &
        \multirow{2}{*}{\makebox[0.15\textwidth][c]{\textbf{Activity-QA}}} &
        \multirow{2}{*}{\makebox[0.1\textwidth][c]{\textbf{Avg.(\%)}}}\\
        \cline{7-8} 
        & & & & & & \textbf{MC} & \textbf{OE} & & \\
        \hline
        \cellcolor{gray!20}LLaVA-OV & \cellcolor{gray!20}100\% & \cellcolor{gray!20}1.00$\times$ & \cellcolor{gray!20}59.3 & \cellcolor{gray!20}37.1 & \cellcolor{gray!20}0.38 & \cellcolor{gray!20}80.9 & \cellcolor{gray!20}52.5 & \cellcolor{gray!20}58.4 & \cellcolor{gray!20}100\\
        
        LLaVA-OV w/ FastV\textsubscript{Opt.Config.} & 50\% & 0.48$\times$ & 58.8 & 36.1 & 0.38 & 80.5 & 51.4 &58.2 & 98.9\\
        \textbf{LLaVA-OV w/ Ours} & 50\% & 0.46$\times$ & 58.8 & 37.2 & 0.38 & 80.2 & 52.0 & 58.1 & \textbf{99.5} \\

        \hline
        LLaVA-OV w/ FastV\textsubscript{Opt.Config.} & 25\% & 0.25$\times$ & 57.0 & 35.4 & 0.35 & 79.7 & 50.9 & 57.2 & 96.2\\
        \textbf{LLaVA-OV w/ Ours} & 25\% & 0.22$\times$ & 56.5 & 36.9 & 0.36 & 79.2 & 51.0 & 58.0 & \textbf{97.3}\\
        
        \hline
        \cellcolor{gray!20}Qwen2-VL & \cellcolor{gray!20}100\% & \cellcolor{gray!20}1.00$\times$ & \cellcolor{gray!20}57.1 & \cellcolor{gray!20}52.7 & \cellcolor{gray!20}1.42 & \cellcolor{gray!20}80.7 & \cellcolor{gray!20}49.5 & \cellcolor{gray!20}57.6 & \cellcolor{gray!20}100\\
        
        Qwen2-VL w/ FastV\textsubscript{Opt.Config.} & 50\% & 0.48$\times$ & 55.4 & 51.3 & 1.40 & 79.6 & 49.0 & 55.7 & 97.9\\
        \textbf{Qwen2-VL w/ Ours} & 50\% & 0.46$\times$ & 55.7 & 51.4 & 1.41 & 79.5 & 48.7 & 56.3 & \textbf{98.2}\\

        \hline
        Qwen2-VL w/ FastV\textsubscript{Opt.Config.} & 25\% & 0.25$\times$ & 53.0 & 49.6 & 1.30 & 78.6 & 47.3 & 52.0 & 93.6\\
        \textbf{Qwen2-VL w/ Ours} & 25\% & 0.22$\times$ & 53.2 & 48.6 & 1.30 & 78.4 & 47.6 & 54.3 & \textbf{94.1}\\

        \hline
        \cellcolor{gray!20}VILA & \cellcolor{gray!20}100\% & \cellcolor{gray!20}1.00$\times$ & \cellcolor{gray!20}48.7 & \cellcolor{gray!20}31.7 & \cellcolor{gray!20}1.30 & \cellcolor{gray!20}70.4 & \cellcolor{gray!20}45.8 & \cellcolor{gray!20}55.2 & \cellcolor{gray!20}100\\
        
        VILA w/ FastV\textsubscript{Opt.Config.} & 50\% & 0.49$\times$ & 48.1 & 31.5 & 1.31 & 70.1 & 46.5 & 55.1 & 100.0\\
        \textbf{VILA w/ Ours} & 50\% & 0.47$\times$ & 48.4 & 34.3 & 1.34 & 70.0 & 47.0 & 56.0 & \textbf{102.4}\\
        \hline
        VILA w/ FastV\textsubscript{Opt.Config.} & 25\% & 0.26$\times$ & 46.3 & 31.8 & 1.26 & 69.6 & 45.6 & 54.6 & 98.3\\
        \textbf{VILA w/ Ours} & 25\% & 0.23$\times$ & 47.4 & 35.0 & 1.29 & 70.0 & 46.7 & 55.7 & \textbf{101.5}\\
        
        \bottomrule
    \end{tabular}}
    \label{tab:video_extend}
\end{table*}

\begin{table*}
\caption{\textbf{Efficiency and performance comparison across different methods on MMStar.} Values marked with * indicate that the retention ratio refers to the average proportion of retained tokens across all LLM layers, due to multi-stage compression in PDrop. For FastV, the same retention ratio corresponds to different FLOPs when compression is applied at different layers (2nd and 4th).}
\vspace{8pt}
\renewcommand{\arraystretch}{1.4}
\setlength{\tabcolsep}{4pt}
\centering
\resizebox{0.6\textwidth}{!}{
% \begin{tabular}{c|c|c|c|c}
\begin{tabular}{
>{\arraybackslash}m{5cm} |
>{\centering\arraybackslash}m{2cm} |
>{\centering\arraybackslash}m{1.5cm} |
>{\centering\arraybackslash}m{2.5cm}}
    \toprule
    \textbf{Method} & \textbf{Retention Ratio} & \textbf{FLOPs(T)} & \textbf{Performance Preservation(\%)}\\
    \hline
    \cellcolor{gray!20}LLaVA-OneVision & \cellcolor{gray!20}100\% & \cellcolor{gray!20}12.9 & \cellcolor{gray!20}100 \\
    
    LLaVA-OneVision w/FastV & 25.0\% & 4.2 & 83.9 \\
    LLaVA-OneVision w/FastV & 25.0\% & 3.5 & 66.3 \\
    
    LLaVA-OneVision w/PDrop & 30.0\%* & 3.8 & \textbf{85.2} \\
    LLaVA-OneVision w/PDrop & 25.4\%* & \underline{3.2} & 80.2 \\
    
    \textbf{LLaVA-OneVision w/Ours} & 25.0\% & \textbf{3.1} & \underline{84.8} \\
    \hline
    \cellcolor{gray!20}Qwen2-VL & \cellcolor{gray!20}100\% & \cellcolor{gray!20}9.6 & \cellcolor{gray!20}100 \\
    Qwen2-VL w/FastV & 25.0\% & \underline{3.1} & \underline{90.0} \\
    \textbf{Qwen2-VL w/Ours} & 25.0\% & \textbf{2.4} & \textbf{91.3} \\
    \bottomrule
\end{tabular}}
 \label{tab:mmstar}
\end{table*}

\begin{table*}
\caption{\textbf{Efficiency and performance comparison across different methods on MVBench.} Values marked with * indicate that the retention ratio is reported from the original paper.}
\vspace{8pt}
\renewcommand{\arraystretch}{1.4}
\setlength{\tabcolsep}{5pt}
\centering
\resizebox{0.6\textwidth}{!}{
% \begin{tabular}{c|c|c|c|c}
        \begin{tabular}{
>{\arraybackslash}m{5cm} |
>{\centering\arraybackslash}m{2cm} |
>{\centering\arraybackslash}m{1.5cm} |
>{\centering\arraybackslash}m{2.5cm}}
    \toprule
    \textbf{Method} & \textbf{Retention Ratio} & \textbf{FLOPs(T)} & \textbf{Performance Preservation(\%)}\\
    \hline
    \cellcolor{gray!20}LLaVA-OneVision & \cellcolor{gray!20}100\% & \cellcolor{gray!20}52.7 & \cellcolor{gray!20}100 \\

    LLaVA-OneVision w/FastVID & 25.0\% & 11.7 & \underline{99.3} \\
    LLaVA-OneVision w/PruneVID & 17.0\%* & 11.9 & 99.1 \\
    LLaVA-OneVision w/FastV & 25.0\% & 16.1 & 95.4 \\
    LLaVA-OneVision w/VisionZip & 25.0\% & \underline{11.7} & 94.4 \\
    \textbf{LLaVA-OneVision w/Ours} & 25.0\% & \textbf{11.7} & \textbf{99.5} \\
    \hline
    \cellcolor{gray!20}Qwen2-VL & \cellcolor{gray!20}100\% & \cellcolor{gray!20}48.4 & \cellcolor{gray!20}100 \\
    
    Qwen2-VL w/FastV & 25.0\% & \underline{14.9} & \textbf{94.1} \\
    \textbf{Qwen2-VL w/Ours} & 25.0\% & \textbf{10.9} & \underline{92.2} \\
    \hline
    \cellcolor{gray!20}VILA1.5 & \cellcolor{gray!20}100\% & \cellcolor{gray!20}27.0 & \cellcolor{gray!20}100 \\
    
    VILA1.5 w/FastV & 25.0\% & \underline{8.2} & \underline{100.3} \\
    \textbf{VILA1.5 w/Ours} & 25.0\% & \textbf{6.3} & \textbf{110.4} \\
    \bottomrule
    \end{tabular}}
 \label{tab:mvbench}
\end{table*}

\begin{table*}
    \caption{\textbf{Efficiency analysis based on Qwen2-VL on MMVet.} We evaluate the inference costs in terms of total inference time, prefilling time, FLOPs, and KV cache memory. KV cache memory is computed with consideration of the Grouped Query Attention (GQA) used in practical inference.}
    \vspace{8pt}
    \renewcommand{\arraystretch}{1.2}
    \setlength{\tabcolsep}{1pt}
    \centering
    \resizebox{1\textwidth}{!}{
    \begin{tabular}{l|c|c|ccccc|c}
        \toprule
        \multirow{2}{*}{\makebox[0.25\textwidth][l]{\textbf{Method}}}&
        \multirow{2}{*}{\parbox[c]{1.5cm}{\centering \textbf{Retention Ratio}}}& 
        \multirow{2}{*}{\makebox[0.15\textwidth][c]{\textbf{FLOPs(\boldsymbol{$\times$})}}} &
        \multirow{2}{*}{\parbox[c]{2.5cm}{\centering \textbf{Total Inference Time}}} &
        \multirow{2}{*}{\parbox[c]{1.5cm}{\centering \textbf{Prefilling Time}}} &
        \multirow{2}{*}{\parbox[c]{2cm}{\centering \textbf{KV Cache}}} &
        \multirow{2}{*}{\parbox[c]{2cm}{\centering \textbf{Total Speedup}}} &
        \multirow{2}{*}{\parbox[c]{2cm}{\centering \textbf{Prefilling Speedup}}} &
        \multirow{2}{*}{\parbox[c]{1.5cm}{\centering \textbf{MMVet}}} \\
        &&&&&&&&\\
        \hline
        \cellcolor{gray!20}Qwen2-VL & \cellcolor{gray!20}100\% & \cellcolor{gray!20}1.00$\times$ & \cellcolor{gray!20}7min58s & \cellcolor{gray!20}1min30s & \cellcolor{gray!20}71.2MB & \cellcolor{gray!20}1.00$\times$ & \cellcolor{gray!20}1.00$\times$ & \cellcolor{gray!20}52.0 \\
        
        Qwen2-VL w/ FastV & 25\% & 0.27$\times$ & 6min50s & 0min56s & 19.7MB & 1.17$\times$ & 1.61$\times$ &33.1\\
        Qwen2-VL w/ PDrop & 25\% & 0.25$\times$ & 6min49s & 0min55s & 18.1MB & 1.17$\times$ & 1.64$\times$ & 47.0\\
        Qwen2-VL w/ Dart & 25\% & 0.30$\times$ & 6min51s & 0min57s & 21.6MB & 1.16$\times$ & 1.58$\times$ & 44.5 \\
        \textbf{Qwen2-VL w/ Ours} & 25\% & \textbf{0.24}\boldsymbol{$\times$} & \textbf{6min50s} & \textbf{0min54s} & \textbf{17.8MB} & \textbf{1.17}\boldsymbol{$\times$} & \textbf{1.67}\boldsymbol{$\times$} & \textbf{50.8}\\
        \bottomrule
    \end{tabular}
    }
    \label{tab:efficiency2}
\end{table*}

\section{The use of Large Language Models(LLMs)}
\label{app:5}
In preparing this manuscript, we used a large language model (LLM, specifically GPT-5-mini) solely as a general-purpose writing and editing assistant. The LLM was employed to improve clarity, grammar, and overall presentation of the text. All technical content, experiments results, and interpretations were generated and verified by the authors. The LLM did not contribute to research ideation, experimental design, data analysis, or the writing of original technical content. The authors take full responsibility for all content presented in this paper.

%% file: iclr2026_conference.bib
@String(CVPR= {IEEE Conf. Comput. Vis. Pattern Recog.})

@String(ICCV= {Int. Conf. Comput. Vis.})

@String(ECCV= {Eur. Conf. Comput. Vis.})

@String(NIPS= {Adv. Neural Inform. Process. Syst.})

@String(ICLR = {Int. Conf. Learn. Represent.})

@String(AAAI = {AAAI})

@String(CVPR  = {CVPR})

@String(ICCV  = {ICCV})

@String(ECCV  = {ECCV})

@String(NIPS  = {NeurIPS})

@String(ICLR  = {ICLR})

@string{ACL = {ACL}}

@string{ICML = {ICML}}

@string{MM = {MM}}

@inproceedings{Grad-SAM,
  author       = {Oren Barkan and
                  Edan Hauon and
                  Avi Caciularu and
                  Ori Katz and
                  others},
  title        = {Grad-SAM: Explaining Transformers via Gradient Self-Attention Maps},
  booktitle    = {CIKM},
  year         = {2021}
}

@inproceedings{fitandprune,
  author       = {Weihao Ye and
                  Qiong Wu and
                  Wenhao Lin and
                  Yiyi Zhou},
  title        = {Fit and Prune: Fast and Training-free Visual Token Pruning for Multi-modal
                  Large Language Models},
  booktitle    = AAAI,
  year         = {2025},
}

@inproceedings{GQA,
  author       = {Drew A. Hudson and
                  Christopher D. Manning},
  title        = {{GQA:} {A} New Dataset for Real-World Visual Reasoning and Compositional
                  Question Answering},
  booktitle    = {CVPR},
  year         = {2019},
}

@inproceedings{MMBench,
  author       = {Yuan Liu and
                  Haodong Duan and
                  Yuanhan Zhang and
                  Bo Li and
                  Others},
  title        = {MMBench: Is Your Multi-modal Model an All-Around Player?},
  booktitle    = {ECCV},
  year         = {2024},
}

@inproceedings{SQA,
  author       = {Pan Lu and
                  Swaroop Mishra and
                  Tanglin Xia and
                  Liang Qiu and
                  Others},
  title        = {Learn to Explain: Multimodal Reasoning via Thought Chains for Science
                  Question Answering},
  booktitle    = NIPS,
  year         = {2022},
}

@inproceedings{TextVQA,
  author       = {Amanpreet Singh and
                  Vivek Natarajan and
                  Meet Shah and
                  Yu Jiang and
                  Others},
  title        = {Towards {VQA} Models That Can Read},
  booktitle    = CVPR,
  year         = {2019},
}

@inproceedings{VizWiz,
  author       = {Danna Gurari and
                  Qing Li and
                  Abigale J. Stangl and
                  Anhong Guo and
                  Others},
  title        = {VizWiz Grand Challenge: Answering Visual Questions From Blind People},
  booktitle    = {CVPR},
  year         = {2018},
}

@article{ocrbench,
  author       = {Yuliang Liu and
                  Zhang Li and
                  Mingxin Huang and
                  Biao Yang and
                  Others},
  title        = {OCRBench: on the hidden mystery of {OCR} in large multimodal models},
  journal      = {Sci. China Inf. Sci.},
  year         = {2024},
}

@inproceedings{chartqa,
  author       = {Ahmed Masry and
                  Do Xuan Long and
                  Jia Qing Tan and
                  Shafiq R. Joty and
                  Enamul Hoque},
  title        = {ChartQA: {A} Benchmark for Question Answering about Charts with Visual
                  and Logical Reasoning},
  booktitle    = {ACL},
  year         = {2022},
}

@inproceedings{docvqa,
  author       = {Minesh Mathew and
                  Dimosthenis Karatzas and
                  C. V. Jawahar},
  title        = {DocVQA: {A} Dataset for {VQA} on Document Images},
  booktitle    = {WACV},
  year         = {2021},
}

@inproceedings{pope,
  author       = {Yifan Li and
                  Yifan Du and
                  Kun Zhou and
                  Jinpeng Wang and
                  Wayne Xin Zhao },
  title        = {Evaluating Object Hallucination in Large Vision-Language Models},
  booktitle    = {EMNLP},
  year         = {2023},
}

@inproceedings{efficientlmm_nips24,
  author       = {Jieneng Chen and
                  Luoxin Ye and
                  Ju He and
                  Zhaoyang Wang and
                  Daniel Khashabi},
  title        = {Efficient Large Multi-modal Models via Visual Context Compression},
  booktitle    = NIPS,
  year         = {2024},
}

@inproceedings{DivPrune,
  author       = {Saeed Ranjbar Alvar and
                  Gursimran Singh and
                  Mohammad Akbari and
                  Yong Zhang},
  title        = {DivPrune: Diversity-based Visual Token Pruning for Large Multimodal
                  Models},
  booktitle    = CVPR,
  year         = {2025},
}

@article{Dart,
  author       = {Zichen Wen and
                  Yifeng Gao and
                  Shaobo Wang and
                  Junyuan Zhang and
                  Qintong Zhang},
  title        = {Stop Looking for Important Tokens in Multimodal Language Models: Duplication
                  Matters More},
  journal      = {EMNLP},
  year         = {2025},
}

@inproceedings{BLIP2,
  author       = {Junnan Li and
                  Dongxu Li and
                  Silvio Savarese and
                  Steven C. H. Hoi},
  title        = {{BLIP-2:} Bootstrapping Language-Image Pre-training with Frozen Image
                  Encoders and Large Language Models},
  booktitle    = ICML,
  year         = {2023}
}

@article{Dynamic-VLM,
  author       = {Han Wang and
                  Yuxiang Nie and
                  Yongjie Ye and
                  Guanyu Deng and
                  others},
  title        = {Dynamic-VLM: Simple Dynamic Visual Token Compression for VideoLLM},
  journal      = {arXiv},
  year         = {2024}
}

@inproceedings{LLAMA-vid,
  author       = {Yanwei Li and
                  Chengyao Wang and
                  Jiaya Jia},
  title        = {LLaMA-VID: An Image is Worth 2 Tokens in Large Language Models},
  booktitle    = ECCV,
  year         = {2024}
}

@inproceedings{CheferGW21,
  author       = {Hila Chefer and
                  Shir Gur and
                  Lior Wolf},
  title        = {Transformer Interpretability Beyond Attention Visualization},
  booktitle    = CVPR,
  pages        = {782--791},
  year         = {2021}
}

@article{llm20_arxiv,
  author       = {Tom B. Brown and
                  Benjamin Mann and
                  Nick Ryder and
                  Melanie Subbiah and
                  others},
  title        = {Language Models are Few-Shot Learners},
  journal      = {arXiv},
  year         = {2020},
}

@article{gpt4_arxiv,
  author       = {OpenAI},
  title        = {{GPT-4} Technical Report},
  journal      = {arXiv},
  year         = {2023},
}

@article{llama_arxiv,
  author       = {Hugo Touvron and
                  Thibaut Lavril and
                  Gautier Izacard and
                  Xavier Martinet and
                  others},
  title        = {LLaMA: Open and Efficient Foundation Language Models},
  journal      = {arXiv},
  year         = {2023},
}

@article{deepseek_arxiv,
  author       = {Xiao Bi and
                  Deli Chen and
                  Guanting Chen and
                  Shanhuang Chen and
                  others},
  title        = {DeepSeek {LLM:} Scaling Open-Source Language Models with Longtermism},
  journal      = {arXiv},
  year         = {2024},
}

@inproceedings{llava_nips,
  author       = {Haotian Liu and
                  Chunyuan Li and
                  Qingyang Wu and
                  Yong Jae Lee},
  title        = {Visual Instruction Tuning},
  booktitle    = NIPS,
  year         = {2023},
}

@article{internvl_arxiv,
  author       = {Zhe Chen and
                  Jiannan Wu and
                  Wenhai Wang and
                  Weijie Su and
                  others},
  title        = {InternVL: Scaling up Vision Foundation Models and Aligning for Generic
                  Visual-Linguistic Tasks},
  journal      = {arXiv},
  year         = {2023},
}

@article{gpt4v_arxiv,
  author       = {Zhe Chen and
                  Weiyun Wang and
                  Hao Tian and
                  Shenglong Ye and
                  others},
  title        = {How Far Are We to GPT-4V? Closing the Gap to Commercial Multimodal
                  Models with Open-Source Suites},
  journal      = {arXiv},
  year         = {2024},
}

@article{qwenvl_arxiv,
  author       = {Jinze Bai and
                  Shuai Bai and
                  Shusheng Yang and
                  Shijie Wang and
                  others},
  title        = {Qwen-VL: {A} Frontier Large Vision-Language Model with Versatile Abilities},
  journal      = {arXiv},
  year         = {2023},
}

@article{gemini_arxiv,
  author       = {Rohan Anil and
                  Sebastian Borgeaud and
                  Yonghui Wu and
                  Jean{-}Baptiste Alayrac and
                  others},
  title        = {Gemini: {A} Family of Highly Capable Multimodal Models},
  journal      = {arXiv},
  year         = {2023},
}

@inproceedings{minigpt4_iclr,
  author       = {Deyao Zhu and
                  Jun Chen and
                  Xiaoqian Shen and
                  Xiang Li and
                  others},
  title        = {MiniGPT-4: Enhancing Vision-Language Understanding with Advanced Large Language Models},
  booktitle    = ICLR,
  year         = {2024},
}

@article{qwen2vl_arxiv,
  author       = {Peng Wang and
                  Shuai Bai and
                  Sinan Tan and
                  Shijie Wang and
                  others},
  title        = {Qwen2-VL: Enhancing Vision-Language Model's Perception of the
                  World at Any Resolution},
  journal      = {arXiv},
  year         = {2024},
}

@article{internlmX2.5_arxiv,
  author       = {Pan Zhang and
                  Xiaoyi Dong and
                  Yuhang Zang and
                  Yuhang Cao and
                  others},
  title        = {InternLM-XComposer-2.5: {A} Versatile Large Vision Language Model
                  Supporting Long-Contextual Input and Output},
  journal      = {arXiv},
  year         = {2024},
}

@inproceedings{tome_iclr,
  author       = {Daniel Bolya and
                  Cheng{-}Yang Fu and
                  Xiaoliang Dai and
                  Peizhao Zhang and
                  others},
  title        = {Token Merging: Your ViT But Faster},
  booktitle    = ICLR,
  year         = {2023},
}

@article{visionzip_arxiv,
  author       = {Senqiao Yang and
                  Yukang Chen and
                  Zhuotao Tian and
                  Chengyao Wang and
                  others},
  title        = {VisionZip: Longer is Better but Not Necessary in Vision Language Models},
  journal      = {arXiv},
  year         = {2024},
}

@inproceedings{tempme_iclr,
  author       = {Leqi Shen and
                  Tianxiang Hao and
                  Tao He and
                  Sicheng Zhao and
                  others},
  title        = {TempMe: Video Temporal Token Merging for Efficient Text-Video Retrieval},
  booktitle    = ICLR,
  year         = {2025},
}

@article{folder_arxiv,
  author       = {Haicheng Wang and
                  Zhemeng Yu and
                  Gabriele Spadaro and
                  Chen Ju and
                  others},
  title        = {{FOLDER:} Accelerating Multi-modal Large Language Models with Enhanced Performance},
  journal      = {arXiv},
  year         = {2025},
}

@inproceedings{fastv_eccv,
  author       = {Liang Chen and
                  Haozhe Zhao and
                  Tianyu Liu and
                  Shuai Bai and
                  others},
  title        = {An Image is Worth 1/2 Tokens After Layer 2: Plug-and-Play Inference
                  Acceleration for Large Vision-Language Models},
  booktitle    = ECCV,
  year         = {2024},
}

@article{sparsevlm_arxiv,
  author       = {Yuan Zhang and
                  Chun{-}Kai Fan and
                  Junpeng Ma and
                  Wenzhao Zheng and
                  others},
  title        = {SparseVLM: Visual Token Sparsification for Efficient Vision-Language Model Inference},
  journal      = {arXiv},
  year         = {2024},
}

@article{smallforlarge_arxiv,
  author       = {Wangbo Zhao and
                  Yizeng Han and
                  Jiasheng Tang and
                  Zhikai Li and
                  others},
  title        = {A Stitch in Time Saves Nine: Small {VLM} is a Precise Guidance for Accelerating Large VLMs},
  journal      = {arXiv},
  year         = {2024},
}

@article{pdrop_arxiv,
  author       = {Long Xing and
                  Qidong Huang and
                  Xiaoyi Dong and
                  Jiajie Lu and
                  others},
  title        = {PyramidDrop: Accelerating Your Large Vision-Language Models via Pyramid Visual Redundancy Reduction},
  journal      = {arXiv},
  year         = {2024},
}

@article{tokencarve_arxiv,
  author       = {Xudong Tan and
                  Peng Ye and
                  Chongjun Tu and
                  Jianjian Cao and
                  others},
  title        = {TokenCarve: Information-Preserving Visual Token Compression in Multimodal
                  Large Language Models},
  journal      = {arXiv},
  year         = {2025},
}

@inproceedings{gae_iccv,
  author       = {Hila Chefer and
                  Shir Gur and
                  Lior Wolf},
  title        = {Generic Attention-model Explainability for Interpreting Bi-Modal and
                  Encoder-Decoder Transformers},
  booktitle    = ICCV,
  year         = {2021},
}

@article{fastvid_arxiv,
  author       = {Leqi Shen and
                  Guoqiang Gong and
                  Tao He and
                  Yifeng Zhang and
                  others},
  title        = {FastVID: Dynamic Density Pruning for Fast Video Large Language Models},
  journal      = {arXiv},
  year         = {2025},
}

@article{prunevid_arxiv,
  author       = {Xiaohu Huang and
                  Hao Zhou and
                  Kai Han},
  title        = {PruneVid: Visual Token Pruning for Efficient Video Large Language Models},
  journal      = {arXiv},
  year         = {2024},
}

@article{deco_arxiv,
  author       = {Linli Yao and
                  Lei Li and
                  Shuhuai Ren and
                  Lean Wang and
                  others},
  title        = {DeCo: Decoupling Token Compression from Semantic Abstraction in Multimodal
                  Large Language Models},
  journal      = {arXiv},
  year         = {2024},
}

@article{llavaov_arxiv,
  author       = {Bo Li and
                  Yuanhan Zhang and
                  Dong Guo and
                  Renrui Zhang and
                  others},
  title        = {LLaVA-OneVision: Easy Visual Task Transfer},
  journal      = {arXiv},
  year         = {2024},
}

@article{vila_arxiv,
  author       = {Zhijian Liu and
                  Ligeng Zhu and
                  Baifeng Shi and
                  Zhuoyang Zhang and
                  others},
  title        = {{NVILA:} Efficient Frontier Visual Language Models},
  journal      = {arXiv},
  year         = {2024},
}

@article{mme,
  author       = {Chaoyou Fu and
                  Peixian Chen and
                  Yunhang Shen and
                  Yulei Qin and
                  others},
  title        = {{MME:} {A} Comprehensive Evaluation Benchmark for Multimodal Large
                  Language Models},
  journal      = {arXiv},
  year         = {2023},
}

@inproceedings{mmstar,
  author       = {Lin Chen and
                  Jinsong Li and
                  Xiaoyi Dong and
                  Pan Zhang and
                  others},
  title        = {Are We on the Right Way for Evaluating Large Vision-Language Models?},
  booktitle    = NIPS,
  year         = {2024},
}

@inproceedings{mmvet,
  author       = {Weihao Yu and
                  Zhengyuan Yang and
                  Linjie Li and
                  Jianfeng Wang and
                  others},
  title        = {MM-Vet: Evaluating Large Multimodal Models for Integrated Capabilities},
  booktitle    = ICML,
  year         = {2024},
}

@article{seedbench,
  author       = {Bohao Li and
                  Rui Wang and
                  Guangzhi Wang and
                  Yuying Ge and
                  others},
  title        = {SEED-Bench: Benchmarking Multimodal LLMs with Generative Comprehension},
  journal      = {arXiv},
  year         = {2023},
}

@article{videomme,
  author       = {Chaoyou Fu and
                  Yuhan Dai and
                  Yondong Luo and
                  Lei Li and
                  others},
  title        = {Video-MME: The First-Ever Comprehensive Evaluation Benchmark of Multi-modal
                  LLMs in Video Analysis},
  journal      = {arXiv},
  year         = {2024},
}

@inproceedings{mvbench,
  author       = {Kunchang Li and
                  Yali Wang and
                  Yinan He and
                  Yizhuo Li and
                  others},
  title        = {MVBench: {A} Comprehensive Multi-modal Video Understanding Benchmark},
  booktitle    = CVPR,
  year         = {2024},
}

@inproceedings{mmbenchvideo,
  author       = {Xinyu Fang and
                  Kangrui Mao and
                  Haodong Duan and
                  Xiangyu Zhao and
                  others},
  title        = {MMBench-Video: {A} Long-Form Multi-Shot Benchmark for Holistic Video Understanding},
  booktitle    = NIPS,
  year         = {2024},
}

@inproceedings{nextqa,
  author       = {Junbin Xiao and
                  Xindi Shang and
                  Angela Yao and
                  Tat{-}Seng Chua},
  title        = {NExT-QA: Next Phase of Question-Answering to Explaining Temporal Actions},
  booktitle    = CVPR,
  year         = {2021},
}

@inproceedings{activityqa,
  author       = {Zhou Yu and
                  Dejing Xu and
                  Jun Yu and
                  Ting Yu and
                  others},
  title        = {ActivityNet-QA: {A} Dataset for Understanding Complex Web Videos via
                  Question Answering},
  booktitle    = AAAI,
  year         = {2019},
}

@inproceedings{flashattn_nips,
  author       = {Tri Dao and
                  Daniel Y. Fu and
                  Stefano Ermon and
                  Atri Rudra and
                  others},
  title        = {FlashAttention: Fast and Memory-Efficient Exact Attention with IO-Awareness},
  booktitle    = NIPS,
  year         = {2022},
}

@inproceedings{vlmevalkit_mm,
  author       = {Haodong Duan and
                  Junming Yang and
                  Yuxuan Qiao and
                  Xinyu Fang and
                  others},
  title        = {VLMEvalKit: An Open-Source ToolKit for Evaluating Large Multi-Modality
                  Models},
  booktitle    = MM,
  year         = {2024},
}

@inproceedings{multihead_acl,
  author       = {Elena Voita and
                  David Talbot and
                  Fedor Moiseev and
                  Rico Sennrich and
                  others},
  title        = {Analyzing Multi-Head Self-Attention: Specialized Heads Do the Heavy
                  Lifting, the Rest Can Be Pruned},
  booktitle    = ACL,
  year         = {2019},
}

@inproceedings{GuWSYXCZ21,
  author       = {Jie Gu and
                  Feng Wang and
                  Qinghui Sun and
                  Zhiquan Ye and
                  others},
  title        = {Exploiting Behavioral Consistence for Universal User Representation},
  booktitle    = AAAI,
  year         = {2021}
}

@inproceedings{kv-cache,
  author       = {Reiner Pope and
                  Sholto Douglas and
                  Aakanksha Chowdhery and
                  Jacob Devlin and
                  others},
  title        = {Efficiently Scaling Transformer Inference},
  booktitle    = {Conf. Mach. Learn. Syst.},
  year         = {2023}
}

@inproceedings{Xception,
  author       = {Fran{\c{c}}ois Chollet},
  title        = {Xception: Deep Learning with Depthwise Separable Convolutions},
  booktitle    = CVPR,
  year         = {2017}
}

@inproceedings{Adam,
  author       = {Diederik P. Kingma and
                  Jimmy Ba},
  title        = {Adam: {A} Method for Stochastic Optimization},
  booktitle    = ICLR,
  year         = {2015}
}

@article{llava-video,
  author       = {Yuanhan Zhang and
                  Jinming Wu and
                  Wei Li and
                  Bo Li and
                  others},
  title        = {Video Instruction Tuning With Synthetic Data},
  journal      = {arXiv},
  year         = {2024}
}

@article{Infinity-MM,
  author       = {Shuhao Gu and
                  Jialing Zhang and
                  Siyuan Zhou and
                  Kevin Yu and
                  others},
  title        = {Infinity-MM: Scaling Multimodal Performance with Large-Scale and High-Quality
                  Instruction Data},
  journal      = {arXiv},
  year         = {2024}
}
